\documentclass[11pt]{article}

\usepackage[table]{xcolor}

\usepackage[final]{acl}

\usepackage{times}
\usepackage{latexsym}
\usepackage[T1]{fontenc}
\usepackage[utf8]{inputenc}
\usepackage{microtype}
\usepackage{inconsolata}
\usepackage{graphicx}

\usepackage{booktabs}
\usepackage{multirow}
\usepackage{colortbl}
\usepackage{pgf} 
\usepackage{enumitem}
\usepackage{tcolorbox}
\tcbuselibrary{skins, breakable}
\usepackage{subcaption}
\usepackage{amsmath}
\usepackage{array}
\usepackage{makecell}
\usepackage{tabularx}
\usepackage{pifont}
\usepackage{hyperref}
\usepackage{ragged2e}
\usepackage{amssymb} 

\newcommand{\boldpara}[1]{%
  \par\vspace{1ex}\noindent\textbf{#1}\space
}
\setlist[itemize]{leftmargin=*}
\setitemize[1]{itemsep=0pt,partopsep=0pt,parsep=\parskip,topsep=0pt}
\DeclareRobustCommand{\logocedar}{%
    \raisebox{-0.15\height}{%
        \includegraphics[height=1em]{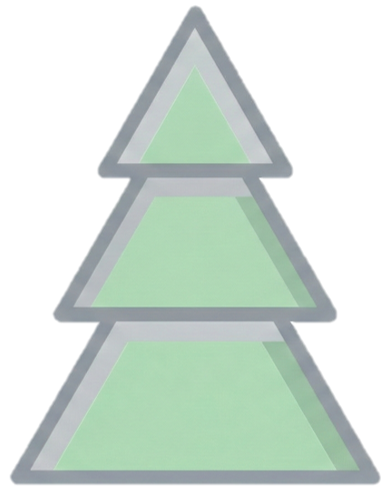}
    }
    \hspace{-0.6em}
}

\newif\ifcomments
\commentstrue  

\definecolor{CedarGray}{RGB}{112, 136, 128}
\newcommand{\Cedar}{\textbf{\textsc{\textcolor{CedarGray}{Cedar}}}}

\definecolor{highcolor}{HTML}{137333}  
\definecolor{lowcolor}{HTML}{FFFFFF}   
\newcommand{\mrbhl}[1]{%
  \pgfmathparse{int(min(100, #1 * 1.6))}
  \xdef\tempcolor{\pgfmathresult}
  \cellcolor{highcolor!\tempcolor!lowcolor}#1
}

\newcommand{\iconReasoning}{
    \raisebox{-0.15\height}{
        \includegraphics[height=1em]{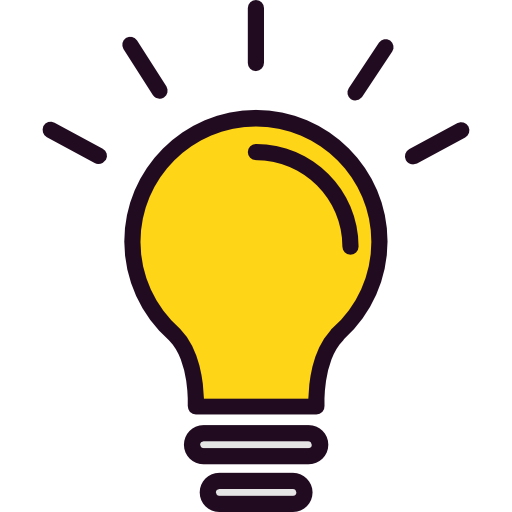}
    }
    \hspace{-0.6em}
}

\newcommand{\iconClosedSource}{
    \raisebox{-0.15\height}{
        \includegraphics[height=1em]{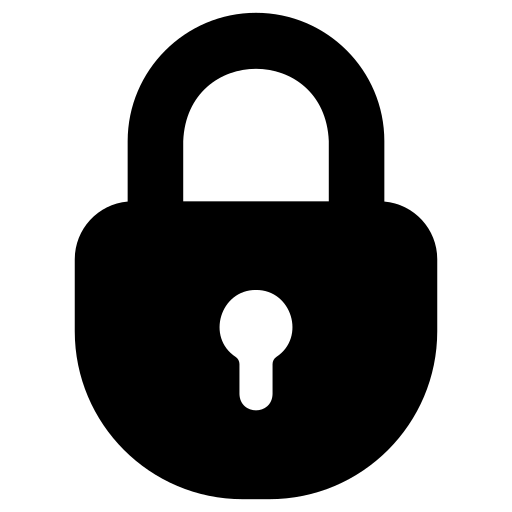}
    }
    \hspace{-1.2em}
}

\newcommand{\iconClosedSourceForgpt}{
    \raisebox{-0.15\height}{
        \includegraphics[height=1em]{images/lock.png}
    }
    \hspace{-0.6em}
}

\newcommand{\iconReasoningForCaption}{%
    \raisebox{-0.15\height}{%
        \includegraphics[height=1em]{images/lightbulb.png}%
    }\hspace{0em}%
}

\newcommand{\iconClosedSourceForCaption}{%
    \raisebox{-0.15\height}{%
        \includegraphics[height=1em]{images/lock.png}%
    }\hspace{0em}%
}

\title{Tears or Cheers? Benchmarking LLMs via Culturally Elicited\\ Distinct Affective Responses}

\author{
\textbf{Chongyuan Dai}$^{1*}$,
\textbf{Yaling Shen}$^{2*}$,
\textbf{Zihan Gao}$^{1}$,
\textbf{Jia Li}$^{1}$,
\textbf{Yishun Jiang}$^{3}$,
\textbf{Yaxiong Wang}$^{1}$,\\
\textbf{Liu Liu}$^{1}$,
\textbf{Zongyuan Ge}$^{2}$,
\textbf{Jinpeng Hu}$^{1\dagger}$\\
    $^1$Hefei University of Technology,
    $^2$Monash University,\\
    $^3$University of Science and Technology of China \\
\texttt{2023217261@mail.hfut.edu.cn, jinpenghu@hfut.edu.cn}
}

\begin{document}
\maketitle

\begingroup
    \renewcommand\thefootnote{}
    \footnotetext{$^*$Equal contribution}
    \footnotetext{$^\dagger$Corresponding author}
\endgroup

\begin{abstract}
Culture serves as a fundamental determinant of human affective processing and profoundly shapes how individuals perceive and interpret emotional stimuli. 
Despite this intrinsic link extant evaluations regarding cultural alignment within large language models primarily prioritize declarative knowledge such as geographical facts or established societal customs. 
These benchmarks remain insufficient to capture the subjective interpretative variance inherent to diverse sociocultural lenses.
To address this limitation, we introduce \logocedar \Cedar, a multimodal benchmark constructed entirely from scenarios capturing \underline{\textsc{C}}ulturally \underline{\textsc{E}}licited \underline{\textsc{D}}istinct \underline{\textsc{A}}ffective \underline{\textsc{R}}esponses.
To construct \textsc{Cedar}, we implement a novel pipeline that leverages LLM-generated provisional labels to isolate instances yielding cross-cultural emotional distinctions, and subsequently derives reliable ground-truth annotations through rigorous human evaluation.
The resulting benchmark comprises 10,962 instances across seven languages and 14 fine-grained emotion categories, with each language including 400 multimodal and 1,166 text-only samples.
Comprehensive evaluations of 17 representative multilingual models reveal a dissociation between language consistency and cultural alignment, demonstrating that culturally grounded affective understanding remains a significant challenge for current models.
Codes and datasets are available at {\url{https://github.com/MindIntLab-HFUT/CEDAR}}. 
\end{abstract}

\section{Introduction}
Culture is the fundamental medium for the construction of human cognition and emotion \cite{kitayama2010handbook}.
The capacity to accurately discern and internalize cultural nuances is indispensable for capturing the latent semantics of natural language and visual signals.
Although recent large language models (LLMs) have demonstrated remarkable proficiency across various domains owing to their text understanding capabilities \cite{zhang-etal-2023-huatuogpt, hu2024psycollm, dai2025psycher1reliablepsychologicalllms, li2025add, shi2026codeocreffectivenessvisionlanguage}, they exhibit uneven cultural understanding, leading to cross-cultural bias and cross-lingual inconsistency.

\begin{figure}[t]
\centering
\includegraphics[width=\columnwidth]{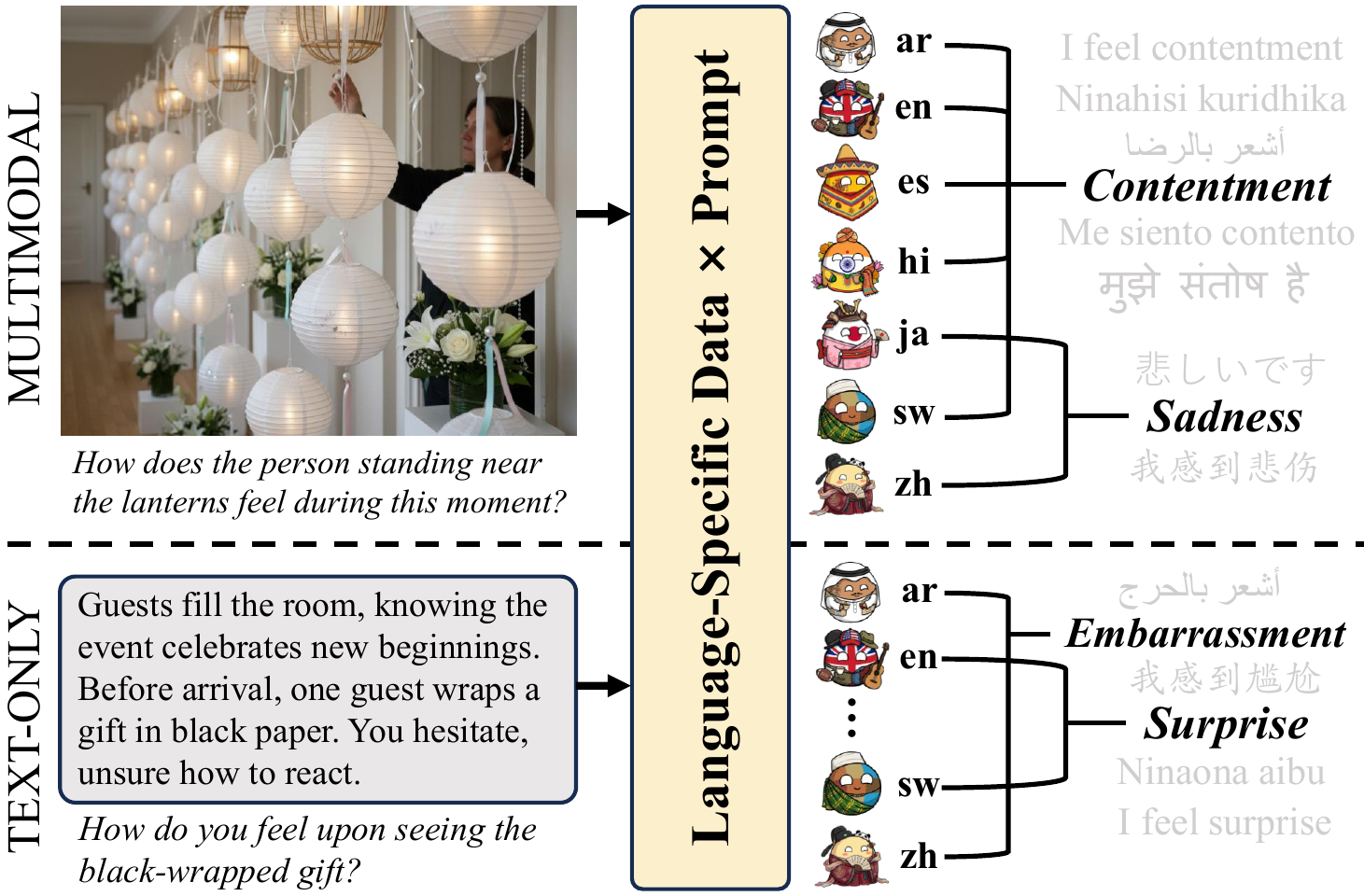}
\caption{
A representative example of culturally distinct scenarios from \textsc{Cedar}.
The figure illustrates how identical multimodal (top) and text-only (bottom) inputs are mapped to different language-specific ground truths.}
\label{fig:intro_case}
\vspace{-1em}
\end{figure}

Therefore, numerous studies have sought to address these disparities by establishing benchmarks to assess cultural commonsense knowledge \cite{li-etal-2024-foodieqa,nayak-etal-2024-benchmarking,zhou-etal-2025-hanfu,onohara-etal-2025-jmmmu} and cultural bias \cite{ramezani-xu-2023-knowledge,dey-etal-2025-llms}.
For instance, CultureAtlas \cite{fung2024massivelymulticulturalknowledgeacquisition} introduces a multicultural dataset derived from Wikipedia, covering a wide range of ethnolinguistic groups.
Similarly, CVQA \cite{romero2024cvqa} curates a benchmark of culturally driven images and questions across languages to assess the cultural awareness of LLMs.
This research trend has spanned diverse linguistic contexts, including benchmarks tailored for Arabic \cite{naous-etal-2024-beer}, Italian \cite{seveso-etal-2025-italic}, and under-represented languages such as Urdu \cite{hashmat-etal-2025-pakbbq} and Southeast Asian languages \cite{satar-etal-2025-seeing}.
However, these studies predominantly focus on declarative cultural commonsense knowledge regarding geography, history, or customs, which overlooks the nuanced dynamics through which specific cultural frameworks modulate the subjective and affective interpretation of information.
Although recent research on cross-cultural emotion understanding has begun to explore this issue \cite{mohamed-etal-2022-artelingo, mohamed2024no, belay-etal-2025-culemo, Hu_Wang_Xie_Li_Ma_Guo_2026}, such efforts are typically confined to narrowly defined domains (e.g., art appreciation) or restricted to a single modality.
In practice, culturally conditioned emotional divergence frequently arises in everyday scenarios.
As illustrated in Figure \ref{fig:intro_case}, a scene featuring white lanterns typically elicits an affective response of mourning within East Asian sociocultural contexts because of their symbolic association with funerary rites.
Conversely, observers from different cultural backgrounds may interpret the same scene as conveying tranquility or even festive contentment.
This example highlights a fundamental challenge in affective understanding: semantic equivalence across languages or visual scenes does not imply emotional equivalence across cultures.
Identical scenarios may elicit substantially different emotional responses depending on the observer’s cultural background.
Nevertheless, most existing benchmarks implicitly assume cultural universality and overlook such culturally grounded variations in emotional interpretation.

Therefore, in this paper, we introduce \logocedar \Cedar, a multimodal benchmark constructed entirely from scenarios capturing \underline{\textsc{C}}ulturally \underline{\textsc{E}}licited \underline{\textsc{D}}istinct \underline{\textsc{A}}ffective \underline{\textsc{R}}esponses.
Unlike prior studies, \textsc{Cedar} exclusively curates instances where the emotional ground truth is contingent upon the cultural context.
This requires LLMs to navigate beyond globalized emotional defaults, thereby assessing their capabilities to align with distinctive cultural frameworks.
Specifically, \textsc{Cedar} comprises 10,962 instances spanning seven languages and 14 fine-grained emotion categories.
To ensure high quality and authentic cultural representation, we implement a novel construction pipeline centered on capturing culturally distinct nuances.
We leverage LLMs to simulate diverse cultural perspectives, thereby identifying scenarios where semantic equivalence fractures into culturally distinct emotional interpretations.
Upon isolating instances with significant cultural distinction, we employ native speakers to rigorously establish the ground-truth labels that reflect authentic cultural perspectives for these candidate instances.
With \textsc{Cedar}, we conduct a comprehensive evaluation of 17 representative multilingual and multimodal LLMs.

Extensive experimental results reveal several noteworthy observations, as detailed below:
\begin{itemize}[leftmargin=*]
\item \textbf{Language modulates emotional distributional shifts.}
For instance, \texttt{Aya-Vision-8B} demonstrates a pronounced inclination to predict \textit{surprise} in \textit{Arabic} contexts, while this pattern is absent in other language groups.
\item \textbf{Systemic prioritization of high arousal states over deactivated emotions.}
Models consistently prioritize high arousal emotions and marginalize deactivated states.
Nevertheless, they exhibit distinct tendencies regarding valence.
While \texttt{Claude4.5-Sonnet} leans towards emotions with high arousal and low valence (10.6\%), \texttt{GPT-4o} shows a contrasting inclination (14.2\%) for high-arousal and pleasant states.
\item \textbf{Dissociation between language consistency and cultural alignment.}
Language consistency does not ensure culturally aligned emotional understanding and may even degrade performance.
For example, \texttt{Gemma3-27B-It} yields only 30.46\% accuracy with Japanese prompts on Japanese data, lagging significantly behind English (44.14\%) and even unrelated Arabic prompts (39.30\%).
\end{itemize}

\section{\logocedar \Cedar: A Benchmark Grounded in Culturally Distinct Emotional Scenarios}

\subsection{Overview}
\begin{figure}[t]
\centering
\includegraphics[width=\columnwidth]{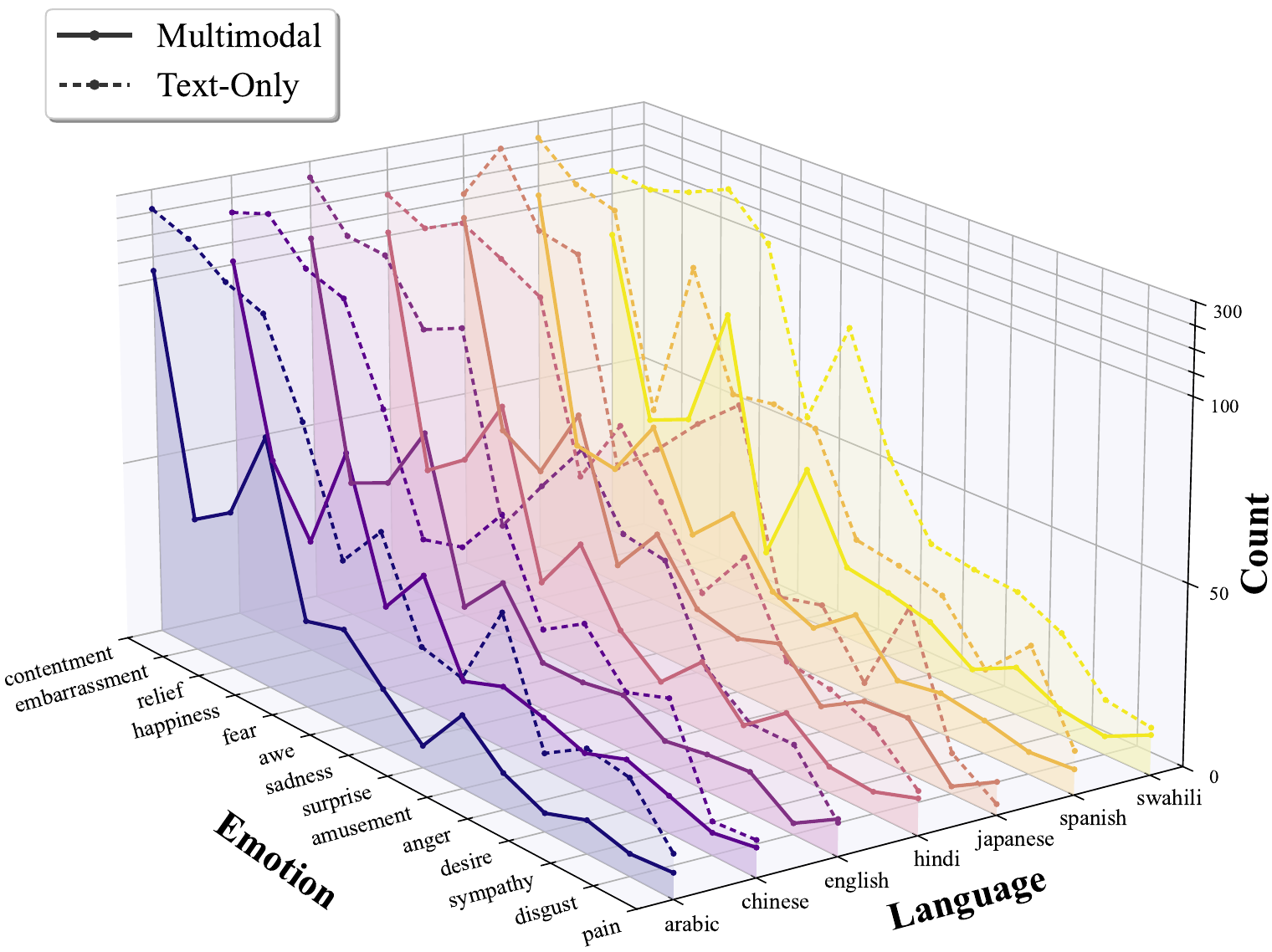}
\caption{
Statistics of \textsc{Cedar}.
}
\label{fig:data_statics}
\vspace{-1em}
\end{figure}

\textsc{Cedar} is a comprehensive benchmark designed to evaluate culturally-grounded emotion alignment in multilingual and multimodal LLMs, comprising scenarios that elicit culturally distinct affective responses.
The benchmark covers seven languages, \textit{Arabic (ar), Chinese (zh), English (en), Hindi (hi), Japanese (ja), Spanish (es), and Swahili (sw)}, and adopts 14 emotion categories adapted from \citet{ekman1992argument} and \citet{cordaro2016voice}.
It contains 10,962 instances, including 400 multimodal and 1,166 text-only samples per language.
We present the statistics of \textsc{Cedar} in Figure \ref{fig:data_statics}.
It demonstrates consistent distributional trends across both modalities, showing a comprehensive coverage of the targeted emotion categories.

In the following subsections, we detail our data curation pipeline.
We start with seed data collection (\S\ref{sec:data_collection}) to generate text-only instances (\S\ref{sec:text_data_construction}).
We subsequently extend these samples to construct multimodal data (\S\ref{sec:multimodal_data_construction}) and conclude with the dataset finalization process (\S\ref{sec:dataset_finalization}).
We present the details of \textsc{Cedar} in Appendix \ref{sec:appendix_details_of_cedar}.

\subsection{Seed Data Collection}
\label{sec:data_collection}
We curate socially grounded seed data from diverse resources, augmented through targeted brainstorming and online retrieval of culturally sensitive contexts.
To ensure language consistency, we employ \texttt{GPT-4.5} to translate the non-English datasets, ArabCulture \cite{sadallah-etal-2025-commonsense} and JETHICS \cite{takeshita2025jethics} into English.
We then standardize all instances into a unified sentence format to address structural disparities among these resources, preparing for subsequent Narrative-Question (NQ) pair generation.

\subsection{Text-Only Data Construction}
\label{sec:text_data_construction}

\boldpara{NQ Pair Generation.}
Building upon the standardized sentences, we prompt \texttt{GPT-4.5} to instantiate each sentence into a NQ pair.
The \textbf{N}arrative establishes a contextual scenario that incorporates situational grounding and participant actions whereas the corresponding \textbf{Q}uestion requires the model to infer the affective state of the protagonist across various temporal stages.
Through this procedure, we obtain approximately 42K candidate NQ pairs covering a wide spectrum of cultural scenarios.

\boldpara{Contextual Refinement.}
To improve textual naturalness and reduce reliance on explicit emotional cues, we apply \texttt{Llama3.3-70B} \cite{grattafiori2024llama} to refine each NQ pair.
Specifically, narratives and questions are rewritten in the second person to enhance subjectivity and immersion. 
Next, phrases containing explicit emotional expressions (e.g., \texttt{``sparking a mix of excitement''}) are removed while ensuring coherence.
This design mitigates information leakage and compels models to infer emotional states from implicit contextual signals rather than overt affective markers.
%

To validate this step, we manually evaluate 100 randomly sampled instances.
Human evaluators confirm that 93\% of the refined narratives successfully removed explicit emotion expressions, while preserving the original situational meaning.

\boldpara{Basic Filtering.}
We initially implement string length constraints to ensure that each instance maintains a character count between 50 and 200. 
This filtering stage accounts for approximately 2.7\% of the candidate pool and guarantees that scenarios provide sufficient context while avoiding unnecessary verbosity.
We then use \texttt{Llama3.3-70B} as a classifier to identify and remove non-social data (e.g., demographic information; $\sim$15.88\%), restricting the benchmark to scenarios that elicit culturally distinct emotional responses.
Finally we utilize \texttt{PolyGuard} \cite{kumar2025polyguardmultilingualsafetymoderation} which is a specialized multilingual safety moderation framework to identify and eliminate toxic content including hate speech or harmful stereotypes. 
By applying these rigorous quality control measures we establish a curated corpus that is both task relevant and safe for broad research applications.

\boldpara{Consistency and Variation Filtering.}
To identify candidate instances likely to elicit culture-specific emotional responses, we employ state-of-the-art LLMs (i.e., \texttt{Claude4.5-Sonnet} \cite{anthropic2025claude45sonnet}, \texttt{Gemini2.5-Flash} \cite{comanici2025gemini}, and \texttt{GPT-4.5}) to generate provisional predictions for each instance across languages.
We first impose \textbf{within-language agreement} by aggregating predictions from the three models for each language. 
Instances exhibiting complete disagreement are discarded as ambiguous, whereas those achieving majority agreement are assigned the corresponding majority label as a provisional emotion. 
We then enforce \textbf{cross-language variation} by comparing these provisional labels across languages and removing instances with uniform predictions, which reflect affective interpretations that are largely invariant across cultural contexts.
This filtering procedure yields approximately 7K NQ pairs that exhibit culturally distinct emotional variation. 
Importantly, the LLM generated provisional labels are used solely for data selection and filtering, and all final ground truth annotations are obtained exclusively through rigorous human labeling.

\subsection{Multimodal Data Construction}
\label{sec:multimodal_data_construction}
In this section, we further extend our pipeline to construct multimodal data in the form of Image-Narrative-Question (INQ) triples.
Each triple integrates an \textbf{I}mage depicting a specific event, a \textbf{N}arrative providing background context, and a \textbf{Q}uestion targeting emotion prediction.

\boldpara{Image Acquisition.}
We employ a hybrid strategy to secure semantically congruent visual stimuli for each Narrative and Question pair. 
We first engage undergraduate and graduate students to retrieve suitable publicly available images via Google or Baidu.
When appropriate visual representations are unavailable, we utilize \texttt{Gemini2.5-Flash-Image} \cite{comanici2025gemini} to generate high fidelity synthetic images that accurately depict the described events.
This dual approach ensures that the visual modality provides a precise grounding for the subsequent affective analysis.

\boldpara{INQ Triple Refinement.}
To prioritize visually grounded reasoning, we prompt \texttt{GPT-4.5} to refine the raw INQ triples via a three-step process:
(1) The model identifies characters and actions as well as interpersonal relationships manifest within the image.
(2) The narrative is rewritten to provide only essential background information such as character backstories that remain distinctively invisible within the image. This process systematically eliminates descriptions of visible content to reduce modal redundancy
(3) The final question is refined to focus on specific characters. This requires models to synthesize the visual scenario with the textual narrative to infer emotional states.

\boldpara{Image Necessity Filtering.}
To ensure the visual modality provides essential information rather than merely illustrating the text, we filter INQ triples based on predictive disparity. 
For each instance, \texttt{GPT-4.5} generates emotion predictions under multimodal and text-only settings; instances with identical predictions are discarded as visually redundant.
This process yields approximately 600 candidate INQ triples, for which provisional labels are generated following the procedure in \S\ref{sec:text_data_construction} to support dataset finalization.

{
\setlength{\tabcolsep}{2.8pt}
\begin{table*}[t]
    \centering
    \resizebox{1\textwidth}{!}{%
        \begin{tabular}{@{}l c | ccccccccc | ccccccccc@{}}
            \toprule
            \multirow{2}{*}{\textbf{Model}} & \multirow{2}{*}{\textbf{Param.}} & \multicolumn{9}{c |}{\textbf{Multimodal}} & \multicolumn{9}{c}{\textbf{Text-only}} \\

            \cmidrule(lr){3-11} \cmidrule(lr){12-20}
             & & AR & EN & ES & HI & JA & SW & ZH & Avg. & Var. & AR & EN & ES & HI & JA & SW & ZH & Avg. & Var. \\
            \midrule

            Aya-Vision-8B & 8B & 
            \mrbhl{31.20} & \mrbhl{42.37} & \mrbhl{30.70} & \mrbhl{33.57} & \mrbhl{30.29} & - & \mrbhl{29.03} & \mrbhl{32.89} & \mrbhl{22.4} & 
            \mrbhl{36.00} & \mrbhl{45.96} & \mrbhl{40.37} & \mrbhl{39.56} & \mrbhl{37.80} & - & \mrbhl{37.43} & \mrbhl{39.58} & \mrbhl{11.9} \\

            \iconReasoning MiniCPM-V-4.5 & 8B & 
            \mrbhl{24.40} & \mrbhl{35.84} & \mrbhl{43.61} & \mrbhl{16.33} & \mrbhl{19.60} & \mrbhl{19.39} & \mrbhl{24.86} & \mrbhl{26.52} & \mrbhl{91.2} & 
            \mrbhl{31.85} & \mrbhl{46.48} & \mrbhl{41.09} & \mrbhl{25.98} & \mrbhl{13.29} & \mrbhl{7.55} & \mrbhl{28.04} & \mrbhl{27.76} & \mrbhl{165.4} \\
            
            \iconReasoning Qwen3-VL-8B & 8B & 
            \mrbhl{30.32} & \mrbhl{41.88} & \mrbhl{39.04} & \mrbhl{27.85} & \mrbhl{26.39} & \mrbhl{14.52} & \mrbhl{27.66} & \mrbhl{31.88} & \mrbhl{85.6} & 
            \mrbhl{39.36} & \mrbhl{49.70} & \mrbhl{42.25} & \mrbhl{37.08} & \mrbhl{28.24} & \mrbhl{18.24} & \mrbhl{33.90} & \mrbhl{35.64} & \mrbhl{94.1} \\

                        Llama3.2-11B-Vision & 11B & 
            - & \mrbhl{34.29} & \mrbhl{28.12} & \mrbhl{32.92} & - & - & - & \mrbhl{31.22} & \mrbhl{10.2} & 
            - & \mrbhl{43.04} & \mrbhl{21.86} & \mrbhl{32.84} & - & - & - & \mrbhl{32.61} & \mrbhl{112.5} \\
            
            Pixtral-12B & 12B & 
            \mrbhl{26.09} & \mrbhl{38.01} & \mrbhl{35.29} & \mrbhl{29.55} & \mrbhl{24.07} & - & \mrbhl{25.44} & \mrbhl{29.95} & \mrbhl{31.4} & 
            \mrbhl{32.65} & \mrbhl{42.37} & \mrbhl{37.27} & \mrbhl{30.28} & \mrbhl{33.16} & - & \mrbhl{31.81} & \mrbhl{35.40} & \mrbhl{19.2} \\
            
            Aya-101 & 13B & 
            - & - & - & - & - & - & - & - & - & 
            \mrbhl{29.33} & \mrbhl{33.45} & \mrbhl{24.59} & \mrbhl{38.29} & \mrbhl{33.11} & \mrbhl{30.38} & \mrbhl{30.62} & \mrbhl{30.54} & \mrbhl{15.3} \\

            Mistral-Small-3.2 & 24B & 
            \mrbhl{30.32} & \mrbhl{47.85} & \mrbhl{48.15} & \mrbhl{32.98} & \mrbhl{27.05} & - & \mrbhl{26.92} & \mrbhl{38.72} & \mrbhl{89.4} & 
            \mrbhl{20.12} & \mrbhl{39.97} & \mrbhl{28.75} & \mrbhl{37.26} & \mrbhl{31.77} & - & \mrbhl{32.96} & \mrbhl{31.18} & \mrbhl{41.8} \\

            Gemma3-27B-It & 27B & 
            \mrbhl{38.56} & \mrbhl{48.37} & \mrbhl{46.65} & \mrbhl{44.38} & \mrbhl{30.46} & \mrbhl{40.00} & \mrbhl{30.00} & \mrbhl{39.82} & \mrbhl{51.3} & 
            \mrbhl{39.13} & \mrbhl{53.91} & \mrbhl{48.80} & \mrbhl{46.85} & \mrbhl{36.03} & \mrbhl{43.11} & \mrbhl{40.85} & \mrbhl{44.31} & \mrbhl{36.4} \\

            Aya-Vision-32B & 32B & 
            \mrbhl{38.74} & \mrbhl{45.16} & \mrbhl{44.17} & \mrbhl{29.41} & \mrbhl{31.01} & - & \mrbhl{43.14} & \mrbhl{39.40} & \mrbhl{42.1} & 
            \mrbhl{41.20} & \mrbhl{52.81} & \mrbhl{44.55} & \mrbhl{46.26} & \mrbhl{36.57} & - & \mrbhl{43.32} & \mrbhl{43.88} & \mrbhl{30.1} \\

            \iconReasoning InternVL3.5-38B & 38B & 
            \mrbhl{28.51} & \mrbhl{37.37} & \mrbhl{37.28} & \mrbhl{32.74} & \mrbhl{23.60} & \mrbhl{15.84} & \mrbhl{30.95} & \mrbhl{30.84} & \mrbhl{58.9} & 
            \mrbhl{36.57} & \mrbhl{22.84} & \mrbhl{34.16} & \mrbhl{35.33} & \mrbhl{22.66} & \mrbhl{11.10} & \mrbhl{33.25} & \mrbhl{27.86} & \mrbhl{77.2} \\

            Qwen2-VL-72B & 72B & 
            \mrbhl{37.76} & \mrbhl{49.36} & \mrbhl{47.61} & \mrbhl{35.68} & \mrbhl{30.57} & \mrbhl{22.98} & \mrbhl{28.68} & \mrbhl{36.10} & \mrbhl{87.4} & 
            \mrbhl{46.43} & \mrbhl{57.16} & \mrbhl{46.68} & \mrbhl{45.96} & \mrbhl{33.51} & \mrbhl{24.12} & \mrbhl{44.27} & \mrbhl{42.59} & \mrbhl{108.2} \\
            
            \iconReasoning Qwen3-VL-235B & 235B & 
            \mrbhl{39.85} & \mrbhl{50.38} & \mrbhl{44.53} & \mrbhl{41.71} & \mrbhl{34.34} & \mrbhl{33.42} & \mrbhl{35.88} & \mrbhl{40.01} & \mrbhl{34.1} & 
            \mrbhl{47.95} & \mrbhl{55.15} & \mrbhl{52.88} & \mrbhl{49.06} & \mrbhl{37.19} & \mrbhl{38.08} & \mrbhl{43.22} & \mrbhl{46.21} & \mrbhl{42.5} \\
            
            Kimi-K2-Instruct & 1T & 
            - & - & - & - & - & - & - & - & - & 
            \mrbhl{44.15} & \mrbhl{56.72} & \mrbhl{49.87} & \mrbhl{46.06} & \mrbhl{39.66} & \mrbhl{41.17} & \mrbhl{40.80} & \mrbhl{45.49} & \mrbhl{36.9} \\
            
            \iconClosedSource \iconReasoning Claude4.5-Sonnet & UNK & 
            \mrbhl{39.90} & \mrbhl{49.62} & \mrbhl{45.45} & \mrbhl{40.82} & \mrbhl{32.99} & \mrbhl{32.15} & \mrbhl{36.71} & \mrbhl{39.66} & \mrbhl{38.9} & 
            \mrbhl{49.69} & \mrbhl{61.79} & \mrbhl{33.59} & \mrbhl{43.95} & \mrbhl{29.60} & \mrbhl{38.30} & \mrbhl{47.51} & \mrbhl{44.09} & \mrbhl{102.4} \\

            \iconClosedSource \iconReasoning Gemini2.5-Flash & UNK & 
            \mrbhl{45.75} & \mrbhl{50.28} & \mrbhl{44.05} & \mrbhl{47.73} & \mrbhl{29.07} & \mrbhl{42.60} & \mrbhl{29.32} & \mrbhl{41.10} & \mrbhl{65.2} & 
            \mrbhl{51.48} & \mrbhl{57.53} & \mrbhl{52.98} & \mrbhl{51.27} & \mrbhl{36.31} & \mrbhl{46.49} & \mrbhl{43.12} & \mrbhl{47.31} & \mrbhl{44.8} \\
            
            \iconClosedSourceForgpt GPT-4o & UNK & 
            \mrbhl{36.93} & \mrbhl{36.34} & \mrbhl{41.60} & \mrbhl{39.14} & \mrbhl{32.50} & \mrbhl{39.37} & \mrbhl{26.01} & \mrbhl{35.97} & \mrbhl{27.8} & 
            \mrbhl{46.56} & \mrbhl{53.69} & \mrbhl{44.55} & \mrbhl{47.45} & \mrbhl{42.97} & \mrbhl{50.43} & \mrbhl{40.34} & \mrbhl{46.57} & \mrbhl{19.6} \\
            
            \iconClosedSource \iconReasoning Qwen3-Omni-Flash & UNK & 
            \mrbhl{39.42} & \mrbhl{40.61} & \mrbhl{46.94} & \mrbhl{41.01} & \mrbhl{33.33} & \mrbhl{25.00} & \mrbhl{32.30} & \mrbhl{37.25} & \mrbhl{51.8} & 
            \mrbhl{38.14} & \mrbhl{52.40} & \mrbhl{46.63} & \mrbhl{43.90} & \mrbhl{29.38} & \mrbhl{9.88} & \mrbhl{36.57} & \mrbhl{36.70} & \mrbhl{185.3} \\

            \bottomrule
        \end{tabular}
    }
    \caption{
    Comparison of different LLMs on \textsc{Cedar}.
    We report the standard accuracy and the cross-lingual variance (Var.) for both subsets (all p-values < 0.01).
    Avg. denotes the macro-average accuracy across languages.
    Darker shades indicate higher numerical values.
    The (\protect\iconReasoningForCaption) denotes reasoning-augmented LLMs, while (\protect\iconClosedSourceForCaption) represents closed-source models, and hyphen (-) indicates the absence of official language support.
    }
    \label{table:main_results}
\vspace{-1em}
\end{table*}
}

\subsection{Dataset Finalization}
\label{sec:dataset_finalization}

\boldpara{Cultural Variation-Based Selection.}
We select representative instances that exhibit maximal cultural variation within clusters of semantically similar scenarios.
Narratives are embedded using \texttt{Qwen3-Embedding-8B} \cite{qwen3embedding} to group semantically related instances. 
Cultural variation within each cluster is quantified leveraging Russell’s Circumplex Model \cite{russell1980circumplex}, which represents emotions along Valence (pleasant vs.\ unpleasant) and Arousal (activation vs.\ deactivation) dimensions. 
The 14 emotion categories are mapped to four quadrants: \textbf{Quadrant I} (high valence, high arousal: \textit{amusement}, \textit{happiness}, \textit{surprise}), \textbf{Quadrant II} (low valence, high arousal: \textit{anger}, \textit{disgust}, \textit{fear}, \textit{pain}), \textbf{Quadrant III} (low valence, low arousal: \textit{embarrassment}, \textit{sadness}), and \textbf{Quadrant IV} (high valence, low arousal: \textit{awe}, \textit{contentment}, \textit{desire}, \textit{relief}, \textit{sympathy}). 
We score instances by cross-language quadrant disagreement and select the highest-scoring instance from each semantic cluster, resulting in 400 INQ triples and 1,166 NQ pairs.
We provide detailed information of the Russell's quadrants in Appendix \ref{sec:appendix_details_of_russells_quadrants}. 

\boldpara{Translation and Human Annotation.}
We first translate the English data into six target languages utilizing \texttt{GPT-4.5}, followed by native-speaker verification via crowdsourcing to ensure fluency and correctness.
%
%
Ground-truth labels are obtained through annotation by native speakers of each target language recruited via Prolific\footnote{\url{https://www.prolific.com/}}, with at least five annotators per instance. If no majority vote is reached, two additional annotators are assigned and the final label is determined by the updated majority.
We present further details of translation process in Appendix \ref{sec:details_of_translation_validation} and annotation process in Appendix \ref{sec:details_of_human_annotation}.
%
%
To capture valid emotional nuances beyond the dominant consensus \cite{fleisig-etal-2023-majority}, we provide an additional multi-label analysis of minority annotations in Appendix \ref{sec:appendix_minority_annotations}.

\section{Methods}

\subsection{Task Definition}
We define culturally-grounded emotion alignment as the task of interpreting scenarios through the cultural lens associated with a target language.
Let $\mathcal{L}$ denote the set of seven target languages and $\mathcal{E}$ the set of 14 emotion categories.
The benchmark consists of a multimodal subset $\mathcal{D}_M$ and a text-only subset $\mathcal{D}_T$.
A multimodal instance is represented as $x = (I, N_l, Q_l, \rho_l)$, where $I$ is the image, and $N_l$, $Q_l$, $\rho_l$ are the narrative, question, and instruction prompt in language $l \in \mathcal{L}$. 
Similarly, a text-only instance is defined as $x = (N_l, Q_l, \rho_l)$.
Formally, given an input $x$, the model needs to predict the affective label $y_l \in \mathcal{E}$.
Notably, while inputs are parallel translations that preserve semantic equivalence, the ground-truth label $y_l$ is culture-specific, reflecting the distinct sociocultural norms associated with the target language $l$.

\subsection{Baselines}
To ensure a comprehensive analysis, we select 17 representative multilingual and multimodal LLMs, spanning various scales from 8B to 1T parameters and including both open-source and proprietary models.
We evaluate various types of models, encompassing general LLMs such as \texttt{GPT-4o} \cite{openai2024gpt4ocard}, reasoning-augmented LLMs like \texttt{Claude4.5-Sonnet}, as well as LLMs optimized for multilingual alignment like \texttt{Aya} series.
Details of evaluation are presented in Appendix \ref{sec:appendix_details_of_evaluation}.

\subsection{Analysis Metrics}
We employ a comprehensive suite of metrics for experimental analysis, ranging from standard accuracy to granular cultural alignment analysis.
\boldpara{Standard Accuracy (SA).} 
This is calculated as the percentage of correct predictions over the total number of questions.

\boldpara{Emotion Prediction Propensity (EPP).} 
We quantify the predictive inclination towards specific fine-grained emotion category $e \in \mathcal{E}$.
For a sample set $\mathcal{S}$, this propensity is defined as the ratio:
\begin{equation}
    \rho_e(\mathcal{S}) = \frac{N_{\mathcal{S}}(\hat{y}=e)}{N_{\mathcal{S}}(y=e)}
\end{equation}
where $\hat{y}$ represents the emotion prediction.

We report \textbf{Global EPP (GEPP)} where $\mathcal{S}$ is the full benchmark, and \textbf{Language-Specific EPP (LSEPP)} where $\mathcal{S}$ corresponds to the subset of a specific language $l$.

\boldpara{Russell's Quadrant Bias (RQB).} 
To analyze affective tendencies beyond discrete labels, we leverage Russell's Circumplex Model \cite{russell1980circumplex} to map emotions onto the continuous dimensions of Valence and Arousal. 
We measure the distributional deviation in model predictions for each quadrant $q$ within a target sample set $\mathcal{S}$:
\begin{equation}
    \beta_q(\mathcal{S}) = \frac{N_{\mathcal{S}}(\hat{y} \in q) - N_{\mathcal{S}}(y \in q)}{\sum_{k=1}^{4} N_{\mathcal{S}}(y \in k)} \times 100\%
\end{equation}
where $N_{\mathcal{S}}(\cdot)$ denotes the sample count within $\mathcal{S}$. 

We compute \textbf{Global RQB (GRQB)} on the entire benchmark and \textbf{Language-Specific RQB (LSRQB)} for each language $l$ to assess cultural alignment at the categorical level.

\begin{figure*}[t]
\centering
\includegraphics[width=0.99\textwidth]{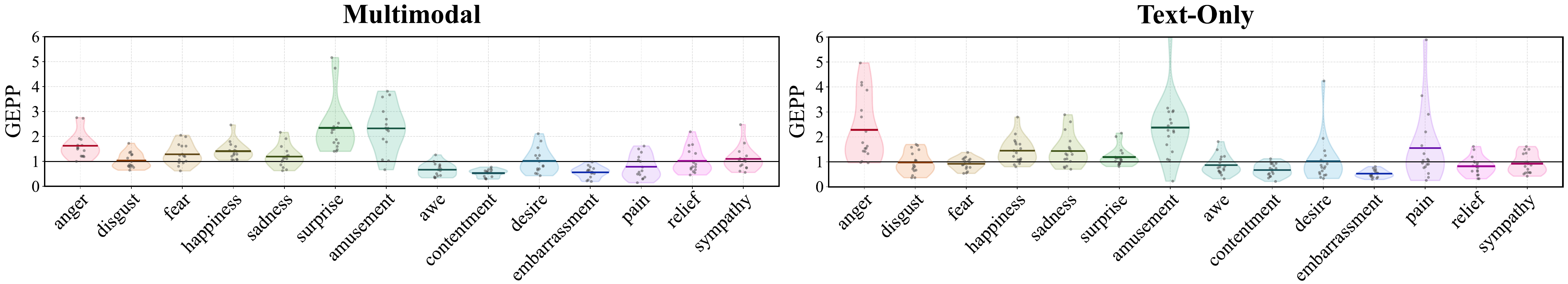} 
\vspace{-1em}
\caption{
Visualization of GEPP across multimodal and text-only subsets.
The scatter points represent the individual performance of the evaluated LLMs for each emotion category.
}
\label{fig:combined_delta_emotion_task1}
\end{figure*}

\begin{figure*}[t]
\centering
\includegraphics[width=0.99\textwidth]{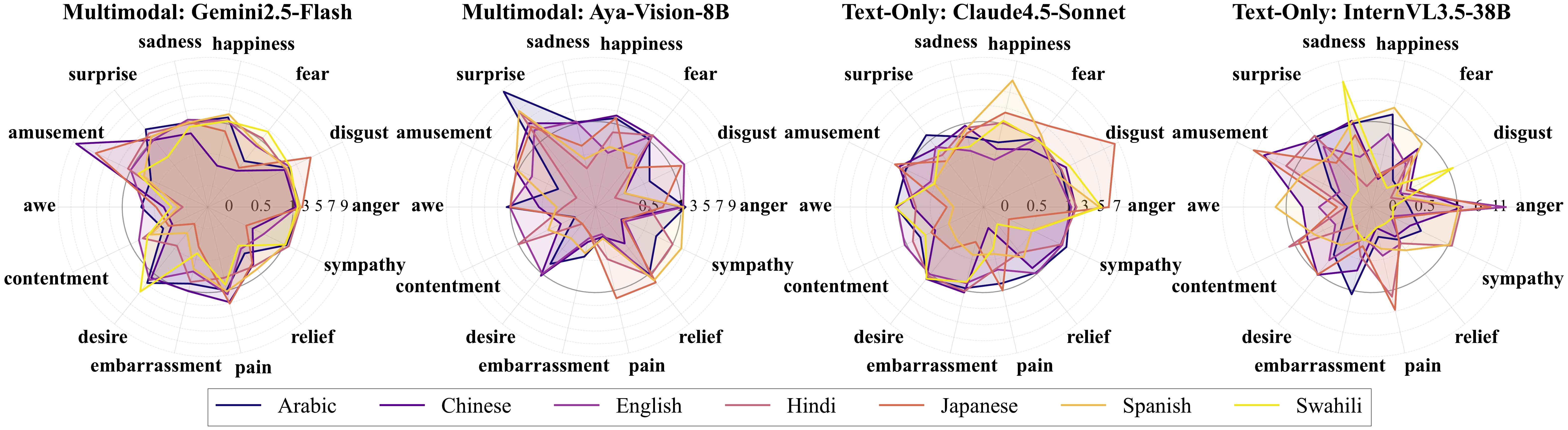}
\caption{
LSEPP for four illustrative models on multimodal and text-only subsets.
Each axis represents the LSEPP value for a specific emotion, demonstrating the variability of bias scores across seven languages.
}
\label{fig:combined_delta_emotion_task2_radar}
\vspace{-1em}
\end{figure*}

\section{Experiments}
\subsection{Overall Performance}
\label{sec:overall_performance}
We present the overall performance of LLMs on \textsc{Cedar} in Table \ref{table:main_results}.
These results reveal several key observations.
First, the evaluated models consistently achieve lower standard accuracy on multimodal instances compared to the text-only subset.
We attribute this to the complexity of visual-emotional grounding in multimodal affective analysis \cite{hu2025beyond, LIAO2026113366}, as interpreting symbolic imagery that carries culturally-grounded emotional weight proves more challenging than processing explicit textual cues.
Second, a performance disparity exists between language groups.
While models consistently perform better in high-resource languages such as \textit{English} and \textit{Spanish}, performance degrades notably in Asian languages (e.g., \textit{Chinese, Japanese}) and low-resource languages like \textit{Swahili}.
This gap indicates that non-Western affective norms are not adequately encoded within these LLMs, emphasizing the critical need for culture-aware emotional alignment.
Third, models explicitly optimized for multilingual alignment (e.g., \texttt{Aya-101} and \texttt{Aya-Vision-8B/32B}) exhibit greater stability across languages.
Despite their smaller model scale, these models retain competitive performance with notably reduced cross-lingual variance.
This finding confirms that targeted optimization for multilingual alignment effectively improves cross-cultural emotional adaptability, underscoring a vital dimension of model development often overlooked in prevailing English-centric research.

\subsection{Emotion Prediction Bias}
\label{sec:emotion_prediction_bias}
To assess the emotion prediction propensities of models, we visualize the GEPP in Figure \ref{fig:combined_delta_emotion_task1}.
In the multimodal subset, models exhibit a distinct propensity towards salient emotions such as \textit{surprise} and \textit{amusement}, while under-predicting subtle states like \textit{contentment} and \textit{embarrassment}.
This disparity suggests that models consistently favor broad emotional categories and frequently fail to capture fine-grained nuances.
In the text-only set, we observe a parallel trend where models tend to over-predict salient emotions at the expense of more complex states.
Notably, these over-predicted emotions are accompanied by markedly higher variance and extreme outliers.
Such instability reflects profound uncertainty in emotional responses across different model series, highlighting current limitations in modeling affective granularity.

We further investigate cross-lingual emotional adaptability by analyzing the LSEPP, with illustrative examples presented in Figure \ref{fig:combined_delta_emotion_task2_radar}.
It illustrates the specific prediction profiles for each model and reveals how these propensities shift according to target languages.
Our results indicate that different language groups exhibit distinct distributions across emotion dimensions. 
For instance, within the \textit{Arabic} group, \texttt{Aya-Vision-8B} demonstrates a pronounced over-prediction of \textit{surprise} in the multimodal subset, whereas this pattern is absent in other language groups.
In addition, we find that models with higher overall accuracy exhibit greater stability and more balanced emotion distributions across languages, suggesting a positive correlation between predictive performance and cross-lingual affective robustness.
For instance, in the multimodal subset, a comparison between \texttt{Gemini2.5-Flash} and \texttt{Aya-Vision-8B} shows that \texttt{Gemini2.5-Flash} demonstrates markedly greater stability in Figure~\ref{fig:combined_delta_emotion_task2_radar}, consistent with its higher overall accuracy of 41.10\%, compared to 32.89\% for \texttt{Aya-Vision-8B}.
Meanwhile, this trend persists in the text-only subset, where \texttt{Claude4.5-Sonnet}, with higher overall accuracy, demonstrates lower prediction volatility compared to \texttt{InternVL3.5-38B}.

\begin{figure}[t]
\centering
\includegraphics[width=\columnwidth]{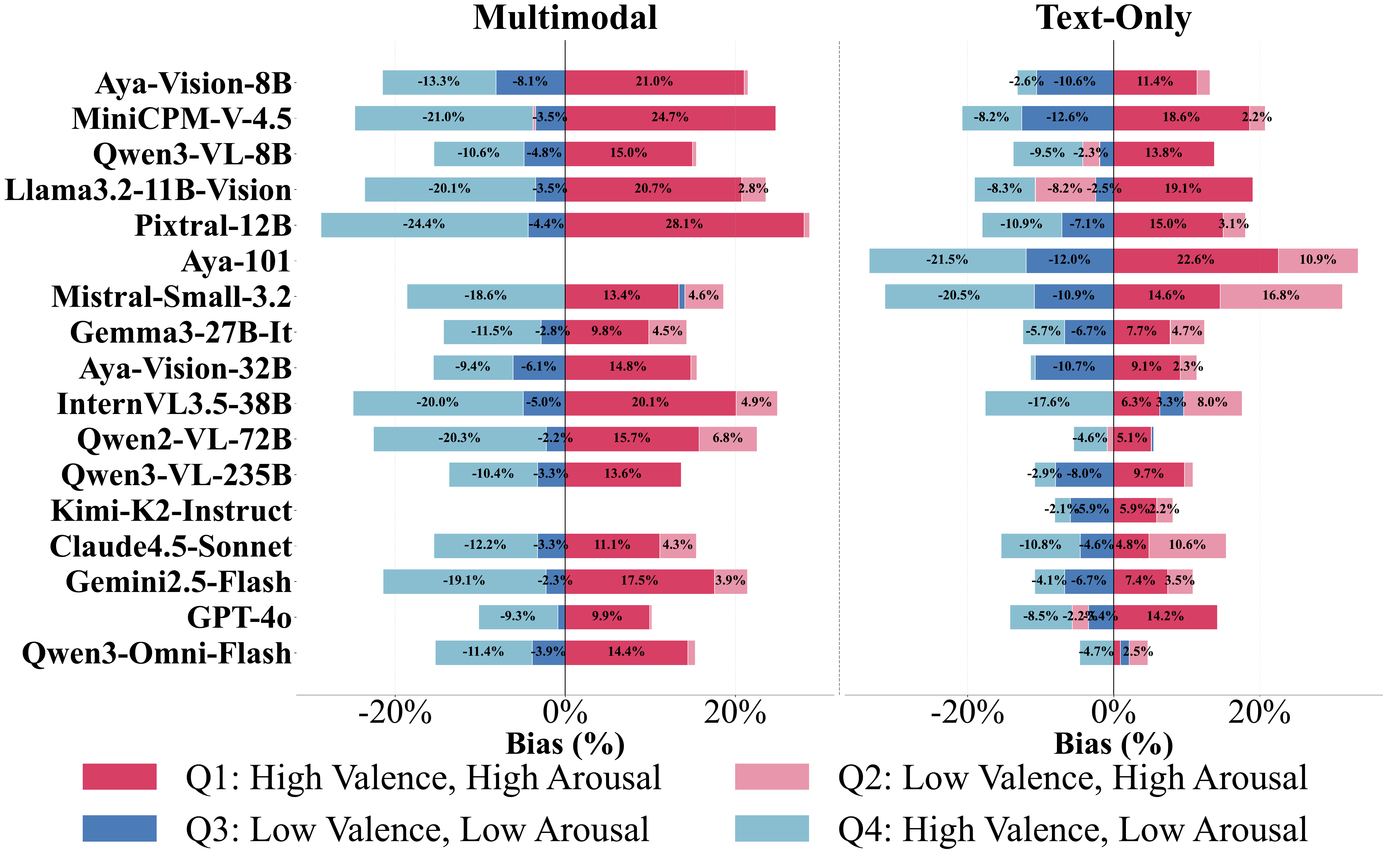}
\caption{
Model comparison based on the GRQB.
The chart shows the GRQB of model predictions across the four quadrants for multimodal and text-only subsets.
}
\label{fig:combined_results_on_russells_model}
\vspace{-1em}
\end{figure}

\subsection{Dimensional Affective Analysis}
\label{sec:Investigating LLMs' Affective Tendencies via Russell's Model}
We adopt Russell’s Circumplex Model \citep{russell1980circumplex}, as described in Section~\ref{sec:dataset_finalization}, to explore models’ affective tendencies, with the corresponding results presented in Figure~\ref{fig:combined_results_on_russells_model}.
At the model level, we observe that LLMs exhibit a systematic preference for high-arousal emotions, while consistently underrepresenting affective states associated with low arousal.
In the multimodal subset, models exhibit a strong and consistent inclination towards high-arousal emotions situated in Quadrant \uppercase\expandafter{\romannumeral1} and Quadrant \uppercase\expandafter{\romannumeral2}, while systematically under-predicting deactivated emotions associated with Quadrants~\uppercase\expandafter{\romannumeral3} and~\uppercase\expandafter{\romannumeral4}.
A similar tendency is observed in the text-only subset, though affective distributions are more diverse.
For instance, \texttt{Claude4.5-Sonnet} and \texttt{Mistral-Small-3.2} prefer low-valence, high-arousal emotions in Quadrant~\uppercase\expandafter{\romannumeral2} (10.6\% and 16.8\%), whereas \texttt{GPT-4o} aligns with Quadrant~\uppercase\expandafter{\romannumeral1} (14.2\%) and \texttt{InternVL3.5-38B} shows a mild bias toward Quadrant~\uppercase\expandafter{\romannumeral3} (3.3\%).
These variations suggest that these models largely share consistent affective tendencies, while exhibiting subtle model-specific differences.

\begin{figure}[t]
\centering
\includegraphics[width=0.96\columnwidth]{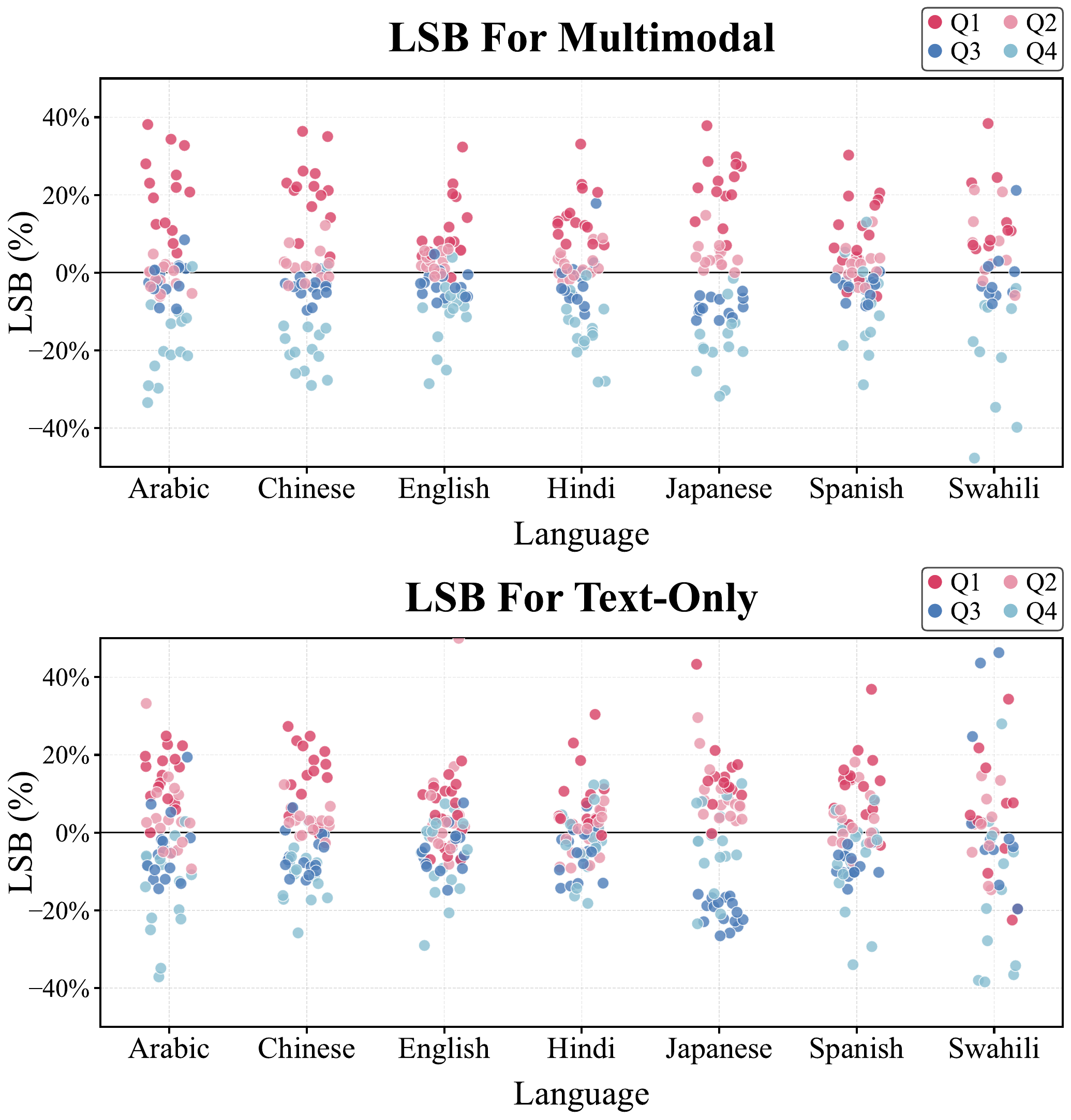}
\caption{
LSRQB values for all evaluated models across languages on multimodal and text-only subsets, categorized by Russell's Quadrants (Q1-Q4).
Each data point represents the result of an individual model.
}
\label{fig:combined_lsb_on_russells_model}
\vspace{-1em}
\end{figure}

We further examine how these affective tendencies shift across languages by analyzing the LSRQB, as shown in Figure \ref{fig:combined_lsb_on_russells_model}.
Across all languages, similar to the trends observed previously evaluated models, models demonstrate a systematic inclination towards emotions in Quadrant \uppercase\expandafter{\romannumeral1} and Quadrant \uppercase\expandafter{\romannumeral2}, while consistently neglecting emotions in Quadrant \uppercase\expandafter{\romannumeral3} and Quadrant \uppercase\expandafter{\romannumeral4}.
Notably, the data points for \textit{English} are tightly clustered around 0\% to exhibit high stability and cross-model consistency, whereas substantially larger dispersion is observed in languages such as \textit{Arabic} and \textit{Swahili}.
The observed increase in volatility indicates that current models exhibit inconsistent generalization across languages, stemming from insufficient cultural and linguistic priors, which hinders accurate affective prediction, particularly in low-resource languages.
Moreover, in the text-only setting, the Japanese group exhibits a pronounced negative clustering of approximately -20\% in low-valence, low-arousal emotions.
This unique pattern remains absent across other language groups which underscores a language specific failure in affective alignment.
Such findings suggest that current models struggle to internalize the deactivated emotional nuances characteristic of the Japanese sociocultural context despite their general linguistic proficiency.

\begin{figure}[t]
\centering
\includegraphics[width=0.97\columnwidth]{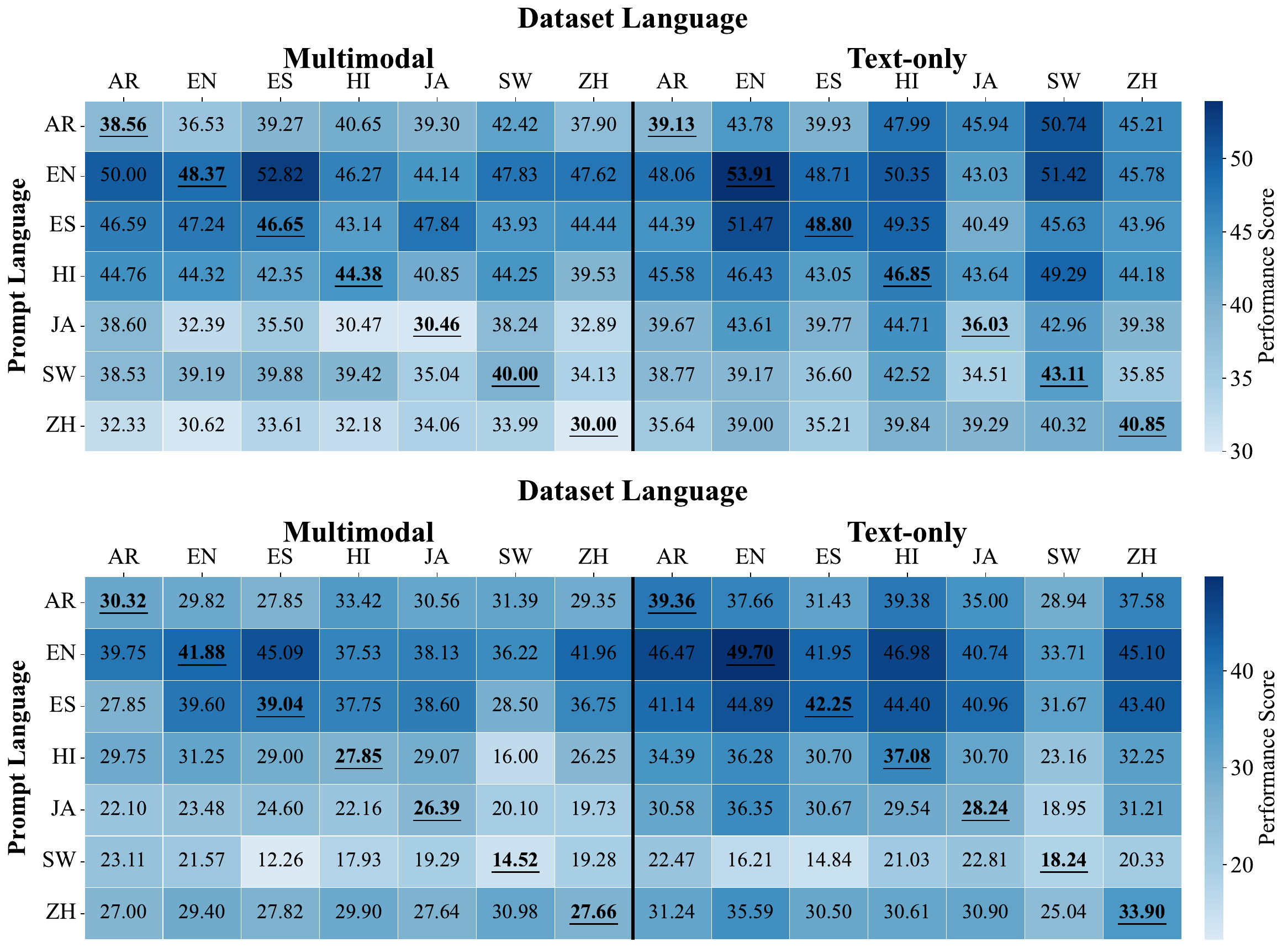}
\caption{
Performance heatmaps analyzing the impact of prompt language ($y$-axis) versus dataset language ($x$-axis) on standard accuracy.
Results are shown for \texttt{Gemma3-27B-It} (top) and \texttt{Qwen3-VL-8B} (bottom) across both multimodal and text-only subsets.
}
\label{fig:combined_data_prompt_heatmap}
\vspace{-1em}
\end{figure}

\subsection{Effect of Prompt Languages}
To investigate the interaction between prompt language and dataset language, we evaluate \texttt{Gemma3-27B-It} and \texttt{Qwen3-VL-8B} under multiple prompt-language configurations, with the results summarized in Figure~\ref{fig:combined_data_prompt_heatmap}.
The results indicate that English prompts consistently yield better performance compared to prompts in other languages.
This advantage is plausibly attributable to the dominance of English data in large-scale pretraining corpora, which induces an inherent English-centric bias in current models.
In contrast, utilizing non-English prompts such as \textit{Arabic} or \textit{Chinese} leads to performance degradation, further emphasizing the critical need for improved cross-lingual consistency and robustness.
Our results reveal a counter intuitive pattern where language consistency between the prompt and the dataset fails to improve performance. 
Although it is commonly assumed that matching the prompt language with the dataset language should be beneficial, our empirical findings show that such alignment fails to yield reliable gains.
For instance, results of \texttt{Gemma3-27B-It} on the multimodal set demonstrate that the combination of a \textit{Japanese} prompt with \textit{Japanese} data yields an accuracy of only 30.46\% which represents the lowest score among comparable groups.
We hypothesize that such language alignment may amplify latent cultural stereotypes or biases associated with the specific language, thereby leading to erroneous emotion predictions.
These findings underscore a profound dissociation between language proficiency and cultural resonance suggesting that models may rely on English as a latent reasoning pivot rather than genuinely internalizing the values associated with native language expressions.

\section{Related Work}
\label{sec:related_work}

\subsection{Cross-Cultural Alignment}
Affective computing has emerged as a frontier in advancing the capabilities of artificial intelligence \cite{song2024emotional,xu-etal-2025-multiagentesc,10.1145/3774904.3792594,yang2025omni}.
Extending this paradigm, recent efforts have been made to develop robust methodologies for cross-cultural alignment \cite{kabir-etal-2025-break}.
A foundational approach involves utilizing LLM-synthesized data to enhance cultural capabilities \cite{li-2024-culturellm, el-mekki-etal-2025-nilechat}.
For example, CARE \cite{guo-etal-2025-care} introduces a preference dataset containing culture-specific instances to improve cultural awareness.
Beyond data augmentation, some studies have proposed specialized learning frameworks for cross-cultural adaptation.
These include contrastive learning approaches designed to capture subtle cultural cues \cite{huang2025cultureclip}, and human-AI collaborative systems engineered to identify and address knowledge gaps \cite{ziems-etal-2025-culture}.
Moreover, recent research has increasingly emphasized pluralism and fairness across cultures.
These efforts range from modular frameworks that leverage multi-LLM collaboration for pluralistic alignment \cite{feng-etal-2024-modular} to analyses of LLM-generated cultural symbols aimed at uncovering and mitigating uneven representational diversity \cite{li2024culturegen}.

\subsection{Cultural Assessments}
There have been numerous studies focusing on the evaluation of cultural commonsense knowledge \cite{nayak-etal-2024-benchmarking, onohara-etal-2025-jmmmu, satar-etal-2025-seeing}.
These encompass assessments of single-nation \cite{seveso-etal-2025-italic}, regional \cite{ma-etal-2025-scalable}, and global-scale cultural contexts \cite{fung2024massivelymulticulturalknowledgeacquisition}.
To capture diverse signals, these efforts have expanded across modalities, including text-only tasks \cite{wang-etal-2024-countries}, visual question answering (VQA; \citet{romero2024cvqa}), video understanding \cite{shafique-etal-2025-culturally}, and text-to-image generation \cite{naous-etal-2024-beer, nayak-etal-2025-culturalframes}.
Furthermore, research has deepened into specific topics such as cuisine \cite{li-etal-2024-foodieqa}, paintings \cite{yu-etal-2025-structured}, and traditional clothing \cite{zhou-etal-2025-hanfu}.
Beyond factual knowledge, parallel works have been proposed to assess cultural paradigms, including moral norms \cite{ramezani-xu-2023-knowledge} and personality traits \cite{dey-etal-2025-llms}.
However, these studies primarily assess static information, overlooking the implicit affective lens through which different cultures interpret identical scenarios.

\section{Conclusion}
In this paper, we introduce \logocedar \Cedar, a multimodal benchmark constructed entirely from scenarios that elicit culturally distinct affective responses.
The benchmark comprises 10,962 instances spanning seven languages and 14 fine grained emotion categories.
To construct this dataset we implement a novel pipeline that utilizes LLMs to simulate diverse cultural perspectives which facilitates the identification of scenarios where identical visual or textual stimuli provoke divergent emotions.
These candidate instances are subsequently validated through rigorous human annotation to ensure definitive ground truth reliability.
Most notably our results expose a dissociation between language proficiency and cultural alignment.
We find that surface level language consistency between the prompt and the dataset does not guarantee the genuine internalization of the underlying sociocultural values required for accurate affective prediction.

\section*{Acknowledgments}

This work was supported in part by National Natural Science Foundation of China under Grant 62402158, Grant 62502145 and Grant 62272144;
by the Key Science \& Technology Project of Anhui Province (202523j08050001);
by the National College Students' Innovation and Entrepreneurship Training Program (202510359110);
by the Anhui Provincial Natural Science Foundation Grant 2408085QF188, and Grant 2408085J040; 
by the Fundamental Research Funds for the Central Universities Grant JZ2025HGTA0162, and Grant JZ2025HGQA0134;
and by the Major Project of Anhui Provincial Science and Technology Breakthrough Program (202423k09020001).

\section*{Limitations}
While \textsc{Cedar} serves as a comprehensive benchmark for cross-cultural emotion alignment, our study is subject to several limitations that remain to be addressed in future research.
\boldpara{Scale and Depth.}
To prioritize high-fidelity cultural validation over extensive language coverage, \textsc{Cedar} focuses on seven representative languages and ensures that every instance undergoes rigorous verification by native speakers.
Limited by resources, we restrict our scope to these cultural clusters to establish a reliable benchmark for assessing the dissociation between language proficiency and cultural alignment.

\boldpara{Affective Theoretical Framework.}
We acknowledge the long-standing discourse within affective science regarding the dichotomy between categorical and dimensional emotion frameworks.
In this paper, we utilize 14 discrete categories to ensure precise quantitative evaluation while actively integrating Russell's Model \cite{russell1980circumplex} for data filtering and dimensional analysis.

\boldpara{Cultural Consensus.}
Constrained by the significant resources required for native-speaker annotations, \textsc{Cedar} strategically targets high-consensus scenarios, validated through strict majority voting, to establish prototypical affective benchmarks.
By focusing on these distinct and widely shared cultural signals, we aim to assess the fundamental capability of LLMs in cross-cultural alignment.

\section*{Ethical Considerations}
\boldpara{Data Safety.}
\textsc{Cedar} is strictly curated to exclude harmful content such as stereotypes or racism.
Beyond implementing a rigorous safety protocol to filter toxic data (\S\ref{sec:text_data_construction}), we emphasize that the ground-truth labels reflect statistical cultural tendencies within a language group, rather than prescriptive stereotypes.
Users should interpret these results as a measure of cultural literacy instead of absolute rules for profiling individuals. 


\bibliography{custom}

\appendix

\section{Details of \logocedar \Cedar}
\label{sec:appendix_details_of_cedar}
\subsection{Comparing \textsc{Cedar} with Cultural Benchmarks}
We compare \textsc{Cedar} with existing benchmarks that focus on cultural commonsense knowledge, cross-cultural bias assessment, and cross-cultural emotion understanding.
Table \ref{table:existing_benchmarks} presents this comparative analysis.
While most existing studies are culture-specific, they either do not include culturally distinct scenarios or only contain such instances partially.
In contrast, \textsc{Cedar} serves as a multimodal benchmark derived entirely from scenarios that evoke culturally distinct emotional responses.
This design allows us to uncover the critical gap between language proficiency and genuine cross-cultural emotional alignment, highlighting how affective patterns expose the internal cultural mechanisms within LLMs.
We present example data from \textsc{Cedar} in Table \ref{tab:text_image_examples} and Table \ref{tab:text_only_examples}.

\subsection{Details of Seed Data}
\label{sec:details_of_seed_data}

To construct a diverse benchmark, we additionally curate seed data from several datasets:

\begin{itemize}[leftmargin=*]
    \item \textbf{ArabCulture} \cite{sadallah-etal-2025-commonsense} is a culturally grounded Arab commonsense dataset comprising 3,482 instances derived from real-world daily life scenarios.

    \item \textbf{Casa} \cite{qiu-etal-2025-evaluating} is a benchmark focusing on social discussion boards and online shopping forums, consisting of 599 entries.

    \item \textbf{Cultural Atlas}\footnote{\url{https://culturalatlas.sbs.com.au/}} is an educational repository detailing the cultural background of migrant populations. We extract 6,304 commonsense knowledge entries spanning 75 countries.

    \item \textbf{CulturalBench} \cite{chiu-etal-2025-culturalbench} is a benchmark designed to assess LLMs' cultural proficiency. It contains 1,227 questions covering 17 diverse topics across 45 global regions, including those underrepresented.

    \item \textbf{CultureBank} \cite{shi-etal-2024-culturebank} is a knowledge base sourced from real-world self-narratives that encapsulate diverse, contextualized cultural scenarios. We retain 16K items with agreement scores exceeding 0.8 to ensure high cultural consensus and plausibility.
    
    \item \textbf{JETHICS} \cite{takeshita2025jethics} is a Japanese dataset presenting various ethical and moral scenarios. We select 14K instances related to commonsense and impartiality, filtering for entries with a Kappa score greater than 0.7.

\end{itemize}

\begin{figure}[t]
\centering
\includegraphics[width=\columnwidth]{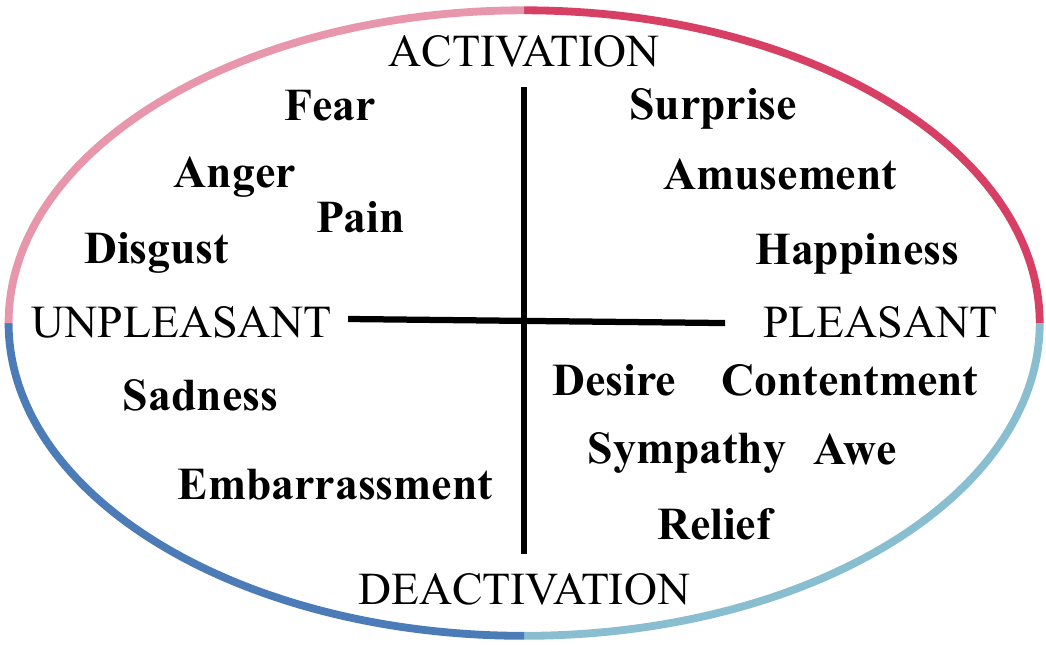}
\caption{
Visualization of the 14 emotion categories on the Russell's Quadrants.
}
\label{fig:russells_quadrants}
\vspace{-1em}
\end{figure}

\subsection{Details of Russell's Quadrants}
\label{sec:appendix_details_of_russells_quadrants}
We map the 14 emotion categories onto Russell's Quadrants as illustrated in Figure \ref{fig:russells_quadrants}.
To represent the continuous nature of affective states, we adapt Russell's Circumplex Model \cite{russell1980circumplex}, which posits that all emotions are distributed in a two-dimensional Euclidean space defined by two orthogonal axes:
Valence (representing the hedonic tone, ranging from unpleasant to pleasant) and Arousal (representing the level of physiological activation, ranging from passive to active).
This dimensional framework facilitates a fine-grained quantification of sentiment, enabling the model to capture subtle gradations in intensity and complex emotional transitions.

\begin{table}[ht]
    \centering
    \small 
    \setlength{\tabcolsep}{6pt}    
    \begin{tabular}{l cc c cc}
        \toprule
        & \multicolumn{2}{c}{\textbf{Multimodal}} && \multicolumn{2}{c}{\textbf{Text-only}} \\
        \cmidrule(lr){2-3} \cmidrule(lr){5-6}
        \textbf{Lang} & $\alpha$ & $\bar{F}_1$ && $\alpha$ & $\bar{F}_1$ \\
        \midrule
        Arabic & 0.732 & 0.579 && 0.792 & 0.673 \\
        Chinese & 0.773 & 0.621 && 0.846 & 0.705 \\
        English & 0.755 & 0.621 && 0.884 & 0.810 \\
        Hindi & 0.741 & 0.715 && 0.878 & 0.650 \\
        Japanese & 0.750 & 0.680 && 0.880 & 0.749 \\
        Spanish & 0.741 & 0.641 && 0.931 & 0.808 \\
        Swahili & 0.693 & 0.577 && 0.925 & 0.806 \\
        \bottomrule
    \end{tabular}
    \caption{Inter-annotator agreement based on Krippendorff's $\alpha$, and Average Pairwise F1-Score ($\bar{F}_1$).}
    \label{tab:agreement_scores}
    \vspace{-1em}
\end{table}

\begin{table*}[t]
  \centering
  \scriptsize
  \setlength{\tabcolsep}{2.0pt}
  \renewcommand{\arraystretch}{0.92}

\newcolumntype{P}[1]{>{\centering\arraybackslash}m{#1}}
\newcolumntype{Z}{>{\centering\arraybackslash}X}
\newcommand{\grp}[1]{\rowcolor{gray!10}\multicolumn{7}{@{}l}{\textit{#1}}\\[2pt]}
\renewcommand\tabularxcolumn[1]{m{#1}} 
\newcommand{\ind}{\hspace*{1.2em}}

\newcommand{\cmark}{\textcolor{green!60!black}{\ding{51}}}
\newcommand{\xmark}{\textcolor{red!70!black}{\ding{55}}}
\newcommand{\benchlink}[2]{\hyperlink{cite.#2}{#1}\nocite{#2}}

\newcommand{\pmark}{
  \raisebox{-0.18em}{
    \includegraphics[height=0.80em]{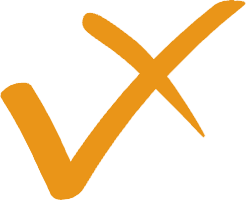}
  }
}

\newcommand{\pmarkforcaption}{%
    \raisebox{-0.18em}{%
        \includegraphics[height=0.80em]{images/halfcheck.png}
    }
    \hspace{-0.6em}
}

  \resizebox{\textwidth}{!}{%
    \begin{tabularx}{\textwidth}{@{}
      >{\raggedright\arraybackslash}m{0.13\textwidth}  
      P{0.065\textwidth}
      P{0.070\textwidth}
      Z
      P{0.13\textwidth}
      P{0.10\textwidth}
      P{0.13\textwidth}
    @{}}
      \toprule
      \rowcolor{gray!15}
      \textbf{Benchmark} &
      \makecell{\textbf{Culture-}\\\textbf{Specific}} &
      \makecell{\textbf{Culturally}\\\textbf{Distinct}} &
      \textbf{Topics} &
      \textbf{Items} &
      \textbf{Languages} &
      \textbf{Modalities} \\
      \midrule

      \grp{Cultural commonsense knowledge}

      \ind \benchlink{CulturalVQA}{nayak-etal-2024-benchmarking}     
                           & \cmark & \xmark & geographically diverse cultural understanding
                           & 2,378
                           & en & image, text \\
      \ind \benchlink{FoodieQA}{li-etal-2024-foodieqa}
                           & \cmark & \xmark & Chinese food culture
                           & 1,364
                           & zh & image, text \\
      \ind \benchlink{Hanfu-Bench}{zhou-etal-2025-hanfu}
                           & \cmark & \xmark & Chinese Hanfu culture
                           & 4,186
                           & zh & image, text \\
      \ind \benchlink{ITALIC}{seveso-etal-2025-italic}
                           & \cmark & \xmark & culture-aware NLU for Italian
                           & 10,000
                           & it & text \\
      \ind \benchlink{JMMMU}{onohara-etal-2025-jmmmu}
                           & \cmark & \xmark & culture-specific evaluation for Japanese
                           & 1,320
                           & ja & image, text \\
      \ind \benchlink{SocialCC}{wu-etal-2025-socialcc}
                           & \cmark & \pmark & interactive cultural competence
                           & 3,060
                           & en & text \\
      \ind \benchlink{Seeing Culture}{satar-etal-2025-seeing}
                           & \cmark & \xmark & two-stage grounding for cultural reasoning
                           & 3,178
                           & en & image, text \\
      \ind \benchlink{ViMUL-Bench}{shafique-etal-2025-culturally}
                           & \cmark & \xmark & culturally diverse multilingual video dataset
                           & 8,025
                           & 14 languages & video, text \\
      \ind \benchlink{CVQA}{romero2024cvqa}
                           & \cmark & \xmark & culturally diverse multilingual VQA
                           & 10,374
                           & 31 languages & image, text \\
      \ind \benchlink{CROPE}{nikandrou-etal-2025-crope}
                           & \cmark & \xmark & in-context adaptation to cultural concepts
                           & 1,060
                           & id, sw, ta, tr, zh & image, text \\

      \cmidrule(lr){1-7}

      \grp{Cross-cultural bias assessment}

      \ind \benchlink{CulturalFrames}{nayak-etal-2025-culturalframes}
                            & \cmark & \xmark & explicit-implicit cultural alignment audit
                            & 3,637
                            & en & image, text \\
      \ind \benchlink{CulturalPersonas}{dey-etal-2025-llms}
                            & \cmark & \pmark & cross-cultural big-five trait alignment
                            & 3,000
                            & en & text \\
      \ind \benchlink{GIMMICK}{schneider-etal-2025-gimmick}
                            & \cmark & \xmark & global cultural-bias benchmarking suite
                            & 7,239
                            & en & video, image, text \\

      \cmidrule(lr){1-7}

      \grp{Cross-cultural emotion understanding}

      \ind \benchlink{ArtELingo}{mohamed-etal-2022-artelingo}
                        & \xmark & \pmark & multilingual art emotion captions
                        & 80,000
                        & en, ar, zh, es & image, text \\
      \ind \benchlink{ArtELingo-28}{mohamed2024no}
                        & \xmark & \pmark & 28-language art emotion captions
                        & 82,000
                        & 28 languages & image, text \\
      \ind \benchlink{CULEMO}{belay-etal-2025-culemo}
                        & \cmark & \pmark & culture-aware emotion prediction
                        & 2,400
                        & 6 languages & text \\

      \midrule
      \logocedar \Cedar & \cmark & \cmark & cultural alignment on culturally distinct scenarios  & 10,962 & 7 languages & image, text \\

      \bottomrule
    \end{tabularx}%
  }

  \caption{Comparison of benchmarks for assessing LLMs' cultural capabilities.
  (\protect\pmarkforcaption) indicates resources that include some instances exhibiting such cultural variation while it is not their primary focus.
  }
  \label{table:existing_benchmarks}
\end{table*}

\begin{table*}[t]
    \centering
    \renewcommand{\arraystretch}{0.95}
    \setlength{\tabcolsep}{3pt}
    
    \scriptsize
    
    \begin{tabular}{@{} l | cccccccc | cccccccc @{}}
        \toprule
        
        \multirow{2}{*}{\textbf{Emotion}} & \multicolumn{8}{c |}{\textbf{Multimodal}} & \multicolumn{8}{c}{\textbf{Text-Only}} \\
        
        \cmidrule(lr){2-9} \cmidrule(lr){10-17}
        
        & ar & zh & en & hi & ja & es & sw & \textbf{Avg.} & ar & zh & en & hi & ja & es & sw & \textbf{Avg.} \\
        
        \midrule

        anger           & 10 & 10 &  8 &  7 &  7 &  9 &  7 & 8.3   & 53 & 45 & 57 & 53 & 35 & 41 & 35 & 45.6 \\
        disgust         &  6 &  6 &  3 &  6 &  2 &  6 &  5 & 4.9   & 26 &  9 & 24 & 23 & 11 & 35 & 15 & 20.4 \\
        fear            & 24 & 23 & 18 & 20 & 20 & 24 & 14 & 20.4  & 79 & 78 & 96 &102 & 48 &101 &123 & 89.6 \\
        happiness       & 70 & 61 & 62 & 65 & 58 & 50 & 78 & 63.4  &132 &131 & 91 &152 &129 & 55 &212 & 128.9 \\
        sadness         & 16 & 13 & 13 & 17 & 18 & 18 & 20 & 16.4  & 59 & 50 & 62 & 74 & 70 & 71 & 88 & 67.7 \\
        surprise        &  6 & 17 & 13 &  8 & 15 & 13 & 18 & 12.9  & 33 & 64 & 77 & 58 & 80 & 69 & 56 & 62.4 \\
        amusement       & 20 & 14 & 15 & 19 & 19 & 22 & 15 & 17.7  & 30 & 38 & 59 & 38 & 32 & 43 & 37 & 39.6 \\
        awe             & 27 & 37 & 30 & 36 & 34 & 35 & 43 & 34.6  & 46 & 47 & 46 & 55 & 58 & 69 & 58 & 54.1 \\
        contentment     &119 &107 &126 &107 &108 &128 & 88 & 111.9 &257 &217 &264 &193 &163 &261 &153 & 215.4 \\
        desire          &  5 & 14 & 10 & 16 & 14 & 11 & 13 & 11.9  & 21 & 32 & 33 & 30 & 19 & 38 & 34 & 29.6 \\
        embarrassment   & 37 & 49 & 38 & 37 & 44 & 35 & 38 & 39.7  &225 &248 &166 &151 &326 &187 &143 & 206.6 \\
        pain            &  7 &  8 & 10 & 10 &  9 &  7 & 11 & 8.9   & 12 & 10 &  9 & 12 &  3 & 12 & 13 & 10.1 \\
        relief          & 44 & 31 & 43 & 45 & 37 & 33 & 43 & 39.4  &165 &161 &158 &197 &147 &161 &171 & 165.7 \\
        sympathy        &  9 & 10 & 11 &  7 & 15 &  9 &  7 & 9.7   & 28 & 36 & 24 & 28 & 45 & 23 & 28 & 30.3 \\

        \bottomrule
    \end{tabular}
    
    \caption{Detailed statistics of the emotion label distribution.}
    \label{table:detailed_datainfo}
\end{table*}

\begin{table*}[t]
    \centering
    \renewcommand{\arraystretch}{0.95}
    
    \small
    
    \begin{tabular*}{0.9\textwidth}{@{\extracolsep{\fill}} l c c c c c c c @{}}
        \toprule
        
        \textbf{Model} & \textbf{Arabic} & \textbf{Chinese} & \textbf{English} & \textbf{Hindi} & \textbf{Japanese} & \textbf{Spanish} & \textbf{Swahili} \\
        
        \midrule
        
        Qwen3-VL-8B      & 0.447 & 0.505 & \textbf{0.668} & 0.413 & 0.391 & 0.570 & 0.096 \\
        Qwen2-VL-72B     & 0.546 & 0.604 & \textbf{0.693} & 0.509 & 0.550 & 0.584 & 0.119 \\
        Claude4.5-Sonnet & 0.618 & 0.608 & \textbf{0.625} & 0.493 & 0.578 & 0.565 & 0.356 \\
        GPT-4o           & 0.666 & 0.583 & \textbf{0.759} & 0.677 & 0.587 & 0.646 & 0.608 \\
        
        \bottomrule
    \end{tabular*}
    
    \caption{Performance comparison across languages under the Top-2-label setting on the text-only subset.}
    \label{tab:minority_results}
\end{table*}

\begin{table*}[t]
    \centering
    \resizebox{0.9\textwidth}{!}{%
        \begin{tabular}{@{}llrr@{}}
            \toprule
            \textbf{Model}  &  \textbf{Param.}  & \textbf{Version}  &  \textbf{Reference}                                 \\
            \midrule
            Aya-101  &  13B  &  CohereLabs/aya-101  &
            \citet{ustun-etal-2024-aya}
            \\
            Aya-Vision-8B  &  8B  &  CohereLabs/aya-vision-8b  &  \citet{dash2025ayavisionadvancingfrontier}
            \\
            Aya-Vision-32B  &  32B  &  CohereLabs/aya-vision-32b  &  \citet{dash2025ayavisionadvancingfrontier}
            \\
            Gemma3-27B-It  &  27B  &  google/gemma-3-27b-it  &  \citet{gemma_2025}
            \\
            \iconReasoning InternVL3.5-38B  &  38B  &  OpenGVLab/InternVL3\_5-38B-HF  &  \citet{wang2025internvl3}
            \\
            Kimi-K2-Instruct  &  1T  &  moonshotai/Kimi-K2-Instruct  &  \citet{kimiteam2026kimik2openagentic}
            \\
            Llama3.2-11B-Vision  &  11B  &  meta-llama/Llama-3.2-11B-Vision-Instruct  &  \citet{grattafiori2024llama}
            \\
            \iconReasoning MiniCPM-V-4.5  &  8B  &  openbmb/MiniCPM-V-4\_5  &  \citet{yu2025minicpm}
            \\
            Mistral-Small-3.2  &  24B  &  mistralai/Mistral-Small-3.2-24B-Instruct-2506  &  \citet{mistralai2025small3}
            \\
            Pixtral-12B  &  12B  &  mistralai/Pixtral-12B-2409  &  \citet{agrawal2024pixtral12b}
            \\
            Qwen2-VL-72B  &  72B  &  Qwen/Qwen2.5-VL-72B-Instruct  &  \citet{wang2024qwen2}
            \\
            \iconReasoning Qwen3-VL-8B  &  8B  &  Qwen/Qwen3-VL-8B-Thinking  &  \citet{qwen3technicalreport}
            \\
            \iconReasoning Qwen3-VL-235B  &  235B  &  Qwen/Qwen3-VL-235B-A22B-Thinking  &  \citet{qwen3technicalreport}
            \\
            \iconClosedSource \iconReasoning Claude4.5-Sonnet  &  UNK  &  claude-sonnet-4-5-20250929  &  \citet{anthropic2025claude45sonnet}
            \\
            \iconClosedSource \iconReasoning Gemini2.5-Flash  &  UNK  &  gemini-2.5-flash-thinking  &  \citet{comanici2025gemini}
            \\
            \iconClosedSourceForgpt GPT-4o  &  UNK  &  gpt-4o-2024-11-20  &  \citet{openai2024gpt4ocard}
            \\
            \iconClosedSource \iconReasoning Qwen3-Omni-Flash  &  UNK  &  Qwen/Qwen3-Omni-Flash-2025-09-15  &  \citet{qwen3technicalreport}
            \\
            \bottomrule
        \end{tabular}
    }
    \caption{Detailed information of the 17 representative multilingual LLMs.}
    \label{table:model_information}
\end{table*}

\subsection{Details of Translation Validation}
\label{sec:details_of_translation_validation}

To assess translation quality, we manually evaluate 100 randomly sampled translated instances.
Native speakers rate each translation on two 10-point Likert scales: factual consistency (FC), which measures the preservation of contextual details, and linguistic fluency, which measures grammatical correctness and naturalness.
The translated data achieves an average score of 9.16/10 for FC and 9.36/10 for fluency, indicating that the translations largely preserve the situational context without introducing substantial artifacts.

\subsection{Details of Human Annotation}
\label{sec:details_of_human_annotation}
To ensure the quality of our ground truths, we additionally recruit two native-speaker annotators per language who are not involved in the
data curation process.
We assign each additional annotator 70 randomly sampled multimodal instances and 100 text-only items for labeling.
We then compare these independent labels with the ground-truth annotations in our initial set.
As shown in Table \ref{tab:agreement_scores}, it demonstrates strong inter-annotator agreements across various metrics, indicating high consistency in the quality judgments.

\subsection{Details of Statistics}
\label{sec:appendix_details_of_statics}
We illustrate the detailed statistics in Table \ref{table:detailed_datainfo}.
We observe that while positive and neutral emotions are prevalent, negative emotions appear less frequently.
Moreover, the distribution of emotion labels also varies among specific language groups.
For instance, in the \textit{Swahili} subset of the multimodal setting, the frequency of \textit{contentment} is only 88 (fall below the average of 111.9), while \textit{happiness} shows a higher prevalence with 78 counts against an average of 63.4.
Similarly, the \textit{Japanese} text-only subset contains 326 instances of \textit{embarrassment} and this figure exceeds the cross-lingual average of 206.6.
These statistical distinctions likely arise from specific cultural norms, demonstrating that our benchmark captures such fine-grained cross-cultural differences.

\section{Analysis on Multi-Label Annotations}
\label{sec:appendix_minority_annotations}

We conduct an additional distribution-aware evaluation on the text-only subset. 
Among the samples, 6,477 instances across the 7 languages exhibit valid minority labels. 
To reflect this distribution, we expand our ground truth to include the Top-2 emotion labels.
We then re-evaluate four representative models using Macro $F_1$, with results presented in Table \ref{tab:minority_results}.

Even under this relaxed, distribution-aware metric, a crucial trend remains: the performance gap across languages persists.
Consistent with the findings in \S\ref{sec:overall_performance}, models continue to achieve higher scores in Western languages (e.g., English, Spanish). 
Conversely, they still exhibit notable performance degradation in Asian languages (e.g., Japanese, Hindi) and low-resource languages like Swahili.
This substantial gap robustly reinforces our core argument: non-Western affective norms remain inadequately encoded within current LLMs, underscoring the necessity of culturally-grounded affective evaluation. 

\section{Details of Evaluation}
\label{sec:appendix_details_of_evaluation}

\begin{figure*}[t]
    \centering
    \begin{minipage}{0.94\textwidth}
        \centering
        \includegraphics[width=\linewidth]{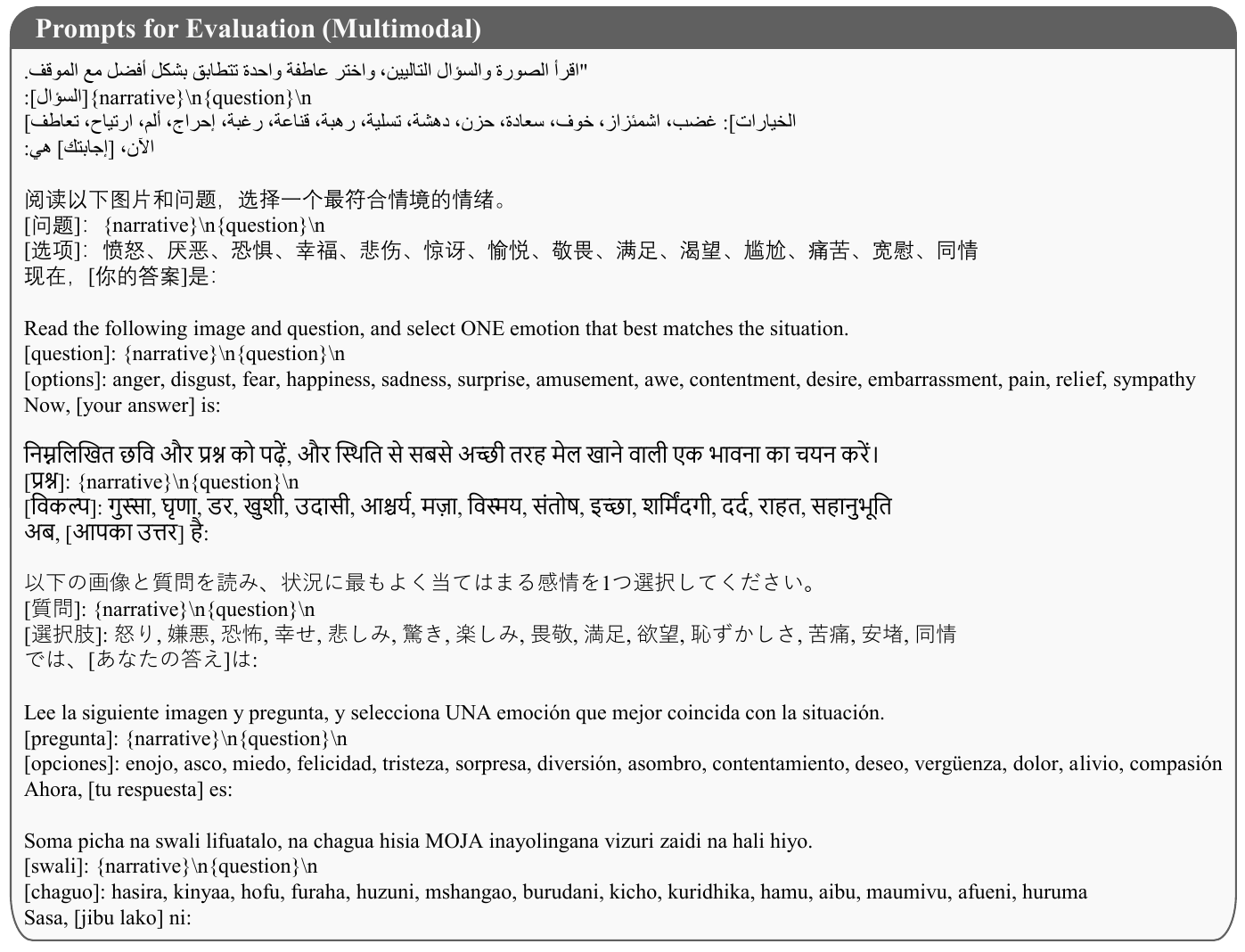}
        \caption{Evaluation prompts for the multimodal subset.}
        \label{fig:evaluation_prompt_multimodal}
    \end{minipage}
    \hfill 
    \begin{minipage}{0.94\textwidth}
        \centering
        \includegraphics[width=\linewidth]{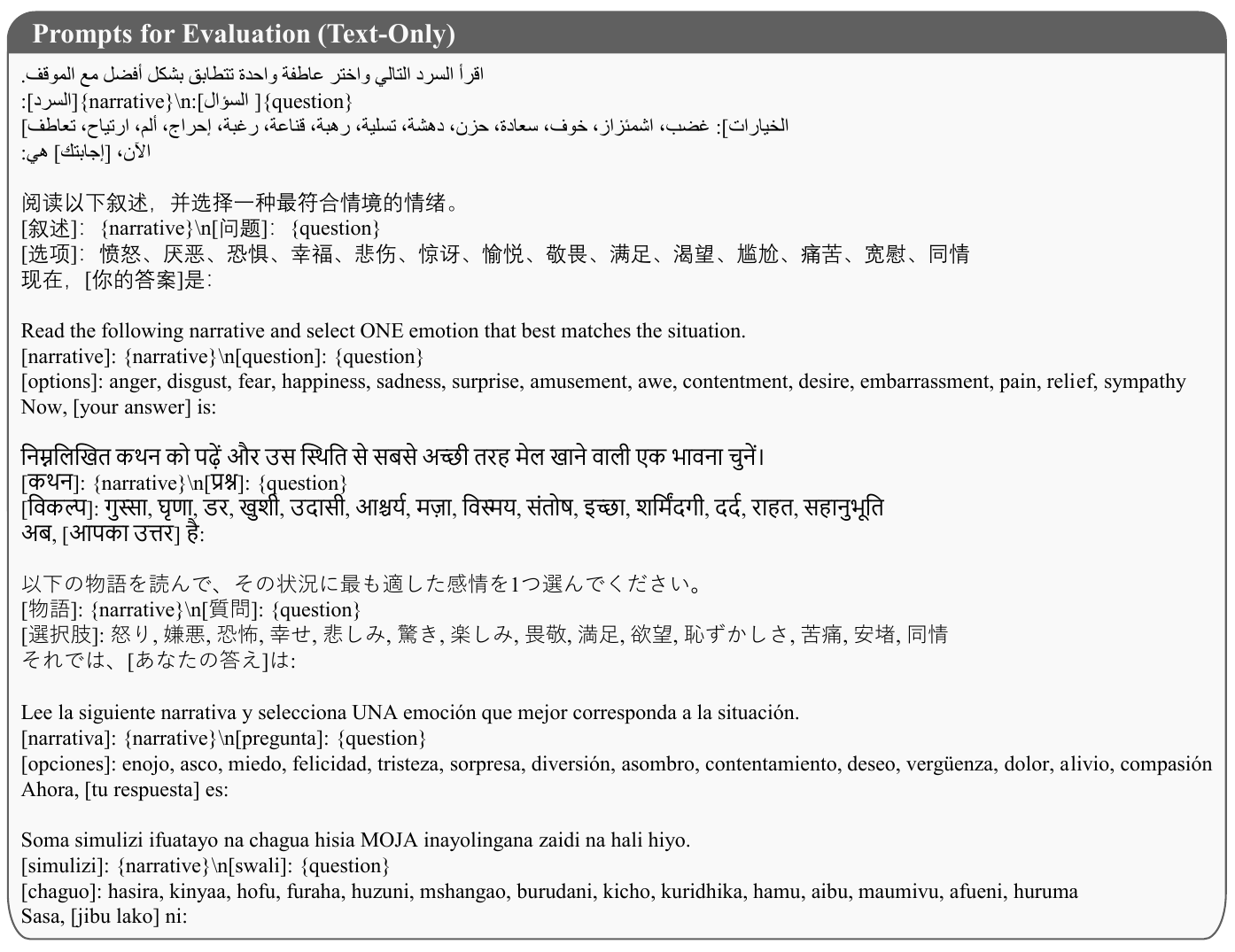}
        \caption{Evaluation prompts for the text-only subset.}
        \label{fig:evaluation_prompt_textonly}
    \end{minipage}
\end{figure*}

We provide a summary of all evaluated LLMs in Table \ref{table:model_information}.
During the evaluation, we set temperature to 0.0, maximum sequence lengths to 128, and top-p to 1.0 to ensure the fairness of evaluation.
Detailed prompts for evaluation are presented in Figure \ref{fig:evaluation_prompt_multimodal} for the multimodal subset and Figure \ref{fig:evaluation_prompt_textonly} for the text-only set.

\setlength{\tabcolsep}{2.8pt}
\begin{table*}[t]
    \centering
    \resizebox{1\textwidth}{!}{%
        \begin{tabular}{@{}l c | c c c c c c c c c c c c c c@{}}
            \toprule
            \multirow{2}{*}{\textbf{Model}} & \multirow{2}{*}{\textbf{Param.}} & \multicolumn{14}{c}{\textbf{Multimodal}} \\
            \cmidrule(lr){3-16}
             & & anger & disgust & fear & happiness & sadness & surprise & amusement & awe & contentment & desire & embarrassment & pain & relief & sympathy \\
            \midrule
            \iconClosedSource \iconReasoning Gemini2.5-Flash & UNK &
            1.91 & 1.73 & 0.94 & 1.45 & 1.13 & 1.47 & 3.00 & 0.35 & 0.58 & 1.55 & 0.72 & 1.62 & 0.69 & 0.87 \\

            \iconReasoning Qwen3-VL-235B & 235B &
            1.19 & 0.65 & 1.25 & 1.21 & 1.04 & 2.37 & 2.33 & 0.51 & 0.66 & 1.24 & 0.66 & 0.47 & 1.39 & 0.56 \\

            Gemma3-27B-It & 27B &
            1.20 & 1.27 & 1.67 & 1.05 & 1.22 & 2.15 & 2.22 & 0.76 & 0.60 & 1.82 & 0.62 & 1.18 & 0.89 & 1.22 \\

            \iconClosedSource \iconReasoning Claude4.5-Sonnet & UNK &
            1.88 & 1.38 & 1.61 & 1.26 & 0.75 & 1.40 & 2.27 & 0.90 & 0.62 & 2.11 & 0.78 & 0.49 & 0.67 & 0.75 \\

            Aya-Vision-32B & 32B &
            1.56 & 1.35 & 1.01 & 1.68 & 0.63 & 2.26 & 1.03 & 1.00 & 0.29 & 0.80 & 0.51 & 0.63 & 2.19 & 1.00 \\

            \iconClosedSourceForgpt GPT-4o & UNK &
            1.49 & 0.82 & 1.20 & 1.06 & 1.16 & 1.73 & 2.48 & 1.26 & 0.56 & 0.83 & 0.85 & 0.30 & 0.96 & 1.74 \\

            Qwen2-VL-72B & 72B &
            2.75 & 0.82 & 2.05 & 1.60 & 1.60 & 1.87 & 1.78 & 0.38 & 0.71 & 0.51 & 0.52 & 0.15 & 0.55 & 0.61 \\

            \iconClosedSource \iconReasoning Qwen3-Omni-Flash & UNK &
            1.21 & 0.87 & 0.96 & 1.29 & 1.11 & 2.29 & 2.34 & 0.70 & 0.77 & 1.00 & 0.57 & 1.37 & 0.79 & 0.82 \\

            Aya-Vision-8B & 8B &
            1.21 & 0.76 & 1.22 & 1.41 & 0.86 & 5.16 & 1.06 & 0.82 & 0.40 & 0.65 & 0.26 & 0.57 & 1.68 & 0.74 \\

            Mistral-Small-3.2 & 24B &
            1.65 & 0.84 & 1.62 & 1.32 & 1.24 & 2.40 & 1.90 & 0.67 & 0.67 & 0.71 & 0.98 & 0.77 & 0.46 & 1.03 \\

            \iconReasoning Qwen3-VL-8B & 8B &
            1.00 & 0.76 & 1.06 & 1.09 & 0.67 & 1.76 & 3.58 & 0.63 & 0.54 & 0.52 & 0.65 & 1.50 & 1.65 & 1.40 \\

            Pixtral-12B & 12B &
            1.43 & 0.80 & 1.29 & 1.80 & 1.00 & 1.60 & 3.82 & 0.87 & 0.31 & 0.43 & 0.57 & 0.31 & 0.78 & 1.12 \\

            Llama3.2-11B-Vision & 11B &
            2.73 & 1.00 & 0.81 & 2.46 & 2.17 & 1.41 & 0.67 & 0.34 & 0.32 & 0.70 & 0.20 & 1.00 & 1.32 & 1.10 \\

            \iconReasoning InternVL3.5-38B & 38B &
            1.53 & 1.28 & 1.99 & 1.44 & 1.41 & 2.54 & 2.69 & 0.44 & 0.56 & 1.24 & 0.37 & 0.37 & 0.59 & 1.05 \\

            \iconReasoning MiniCPM-V-4.5 & 8B &
            1.65 & 1.13 & 0.62 & 1.07 & 1.91 & 4.74 & 3.67 & 0.42 & 0.39 & 1.13 & 0.22 & 1.05 & 0.74 & 2.48 \\

            \bottomrule
        \end{tabular}
    }
    \caption{Detailed emotion prediction bias (\S\ref{sec:emotion_prediction_bias}) on the multimodal subset, evaluated by the GEPP metric.}
    \label{table:appendix_emotion_bias_1}
\end{table*}
\setlength{\tabcolsep}{2.8pt}
\begin{table*}[t]
    \centering
    \resizebox{1\textwidth}{!}{%
        \begin{tabular}{@{}l c | c c c c c c c c c c c c c c@{}}
            \toprule
            \multirow{2}{*}{\textbf{Model}} & \multirow{2}{*}{\textbf{Param.}} & \multicolumn{14}{c}{\textbf{Text-Only}} \\
            \cmidrule(lr){3-16}
             & & anger & disgust & fear & happiness & sadness & surprise & amusement & awe & contentment & desire & embarrassment & pain & relief & sympathy \\
            \midrule

            \iconClosedSource \iconReasoning Gemini2.5-Flash & UNK &
            1.43 & 1.02 & 1.11 & 0.93 & 0.76 & 1.16 & 3.01 & 0.47 & 0.93 & 1.94 & 0.71 & 2.20 & 0.87 & 0.58 \\

            \iconReasoning Qwen3-VL-235B & 235B &
            1.33 & 0.82 & 1.05 & 1.13 & 1.30 & 1.46 & 2.70 & 0.73 & 0.84 & 0.92 & 0.45 & 0.80 & 1.20 & 0.51 \\

            Kimi-K2-Instruct & 1T &
            1.54 & 0.87 & 1.03 & 1.08 & 1.02 & 1.15 & 2.24 & 1.22 & 0.92 & 0.86 & 0.66 & 1.06 & 0.84 & 1.35 \\

            Gemma3-27B-It & 27B &
            1.42 & 1.05 & 1.11 & 1.08 & 1.32 & 1.35 & 2.44 & 0.97 & 0.94 & 1.45 & 0.52 & 3.65 & 0.61 & 0.98 \\

            \iconClosedSource \iconReasoning Claude4.5-Sonnet & UNK &
            2.80 & 1.32 & 1.38 & 1.53 & 0.79 & 0.81 & 1.07 & 0.76 & 0.74 & 1.25 & 0.81 & 0.53 & 0.63 & 0.82 \\

            Aya-Vision-32B & 32B &
            3.06 & 0.78 & 1.19 & 1.45 & 0.70 & 1.15 & 1.40 & 0.63 & 0.41 & 1.07 & 0.45 & 0.91 & 1.62 & 0.87 \\

            \iconClosedSourceForgpt GPT-4o & UNK &
            0.98 & 0.64 & 0.89 & 1.85 & 1.52 & 1.09 & 2.26 & 1.10 & 0.60 & 0.81 & 0.64 & 0.25 & 0.81 & 1.61 \\

            Qwen2-VL-72B & 72B &
            1.63 & 0.36 & 0.74 & 1.22 & 1.83 & 1.07 & 1.69 & 0.67 & 1.12 & 0.64 & 0.75 & 0.77 & 0.77 & 0.57 \\

            \iconClosedSource \iconReasoning Qwen3-Omni-Flash & UNK &
            1.54 & 1.66 & 0.88 & 0.80 & 2.61 & 0.93 & 2.03 & 0.32 & 1.01 & 0.74 & 0.55 & 1.33 & 0.94 & 0.93 \\

            Aya-Vision-8B & 8B &
            2.22 & 0.71 & 0.95 & 1.37 & 1.10 & 2.15 & 1.10 & 1.49 & 0.50 & 0.46 & 0.40 & 0.90 & 1.49 & 0.57 \\

            Mistral-Small-3.2 & 24B &
            4.96 & 1.07 & 1.15 & 2.12 & 1.28 & 0.93 & 2.19 & 1.20 & 0.53 & 0.33 & 0.37 & 0.38 & 0.33 & 0.42 \\

            \iconReasoning Qwen3-VL-8B & 8B &
            0.96 & 0.67 & 0.60 & 1.71 & 2.29 & 0.91 & 3.01 & 0.58 & 0.69 & 0.52 & 0.46 & 2.90 & 1.01 & 0.70 \\

            Pixtral-12B & 12B &
            3.88 & 1.58 & 0.93 & 1.71 & 1.42 & 0.88 & 2.55 & 0.63 & 0.35 & 0.48 & 0.39 & 1.11 & 0.98 & 0.99 \\

            Llama3.2-11B-Vision & 11B &
            1.02 & 1.70 & 0.55 & 1.79 & 1.58 & 1.19 & 3.05 & 1.81 & 0.42 & 0.78 & 0.54 & 5.89 & 0.48 & 1.50 \\

            Aya-101 & 13B &
            4.19 & 0.45 & 0.80 & 2.79 & 1.12 & 2.01 & 0.23 & 0.56 & 0.21 & 0.38 & 0.29 & 1.65 & 0.66 & 1.50 \\

            \iconReasoning InternVL3.5-38B & 38B &
            4.08 & 0.36 & 0.54 & 0.89 & 2.89 & 1.02 & 3.16 & 0.82 & 0.74 & 0.58 & 0.56 & 0.95 & 0.32 & 0.56 \\

            \iconReasoning MiniCPM-V-4.5 & 8B &
            1.77 & 1.49 & 0.78 & 1.08 & 0.83 & 0.99 & 6.21 & 0.80 & 0.53 & 4.24 & 0.34 & 1.06 & 0.47 & 1.32 \\      

            \bottomrule
        \end{tabular}
    }
    \caption{Detailed emotion prediction bias (\S\ref{sec:emotion_prediction_bias}) on the text-only set calculated by the GEPP.}
    \label{table:appendix_emotion_bias_2}
\end{table*}

\begin{table*}[t]
\centering
\small
\renewcommand{\arraystretch}{1.35}
\setlength{\tabcolsep}{3pt}

\newcommand{\Ltag}[1]{\textcolor{black}{\textbf{\texttt{#1}}}}

\newcommand{\AnsBlock}[7]{%
  \parbox[t]{\linewidth}{%
    \vspace{-1.1em} %
    \setlength{\tabcolsep}{2pt}%
    \begin{tabular}[t]{@{}%
      >{\raggedright\arraybackslash}p{0.32\linewidth}%
      >{\raggedright\arraybackslash}p{0.32\linewidth}%
      >{\raggedright\arraybackslash}p{0.32\linewidth}@{}}
      \Ltag{AR}: #1 & \Ltag{ZH}: #2 & \Ltag{EN}: #3 \\
      \Ltag{HI}: #4 & \Ltag{JA}: #5 & \Ltag{ES}: #6 \\
      \Ltag{SW}: #7 &                &                \\
    \end{tabular}%
  }%
}

\newcommand{\ImageCell}[1]{%
  \fbox{%
    \parbox[t][2.6cm][c]{\dimexpr\linewidth-2\fboxsep-2\fboxrule\relax}{%
      \centering
      \includegraphics[width=\linewidth,height=2.6cm,keepaspectratio]{#1}%
    }%
  }%
}

\begin{tabular}{@{}%
  p{0.01\textwidth}%
  p{0.18\textwidth}%
  p{0.31\textwidth}%
  p{0.46\textwidth}%
@{}}

\toprule
\textbf{\#} & \textbf{Image} & \textbf{Narrative \& Question} & \textbf{Multilingual Answers} \\ 
\midrule

1 &
\ImageCell{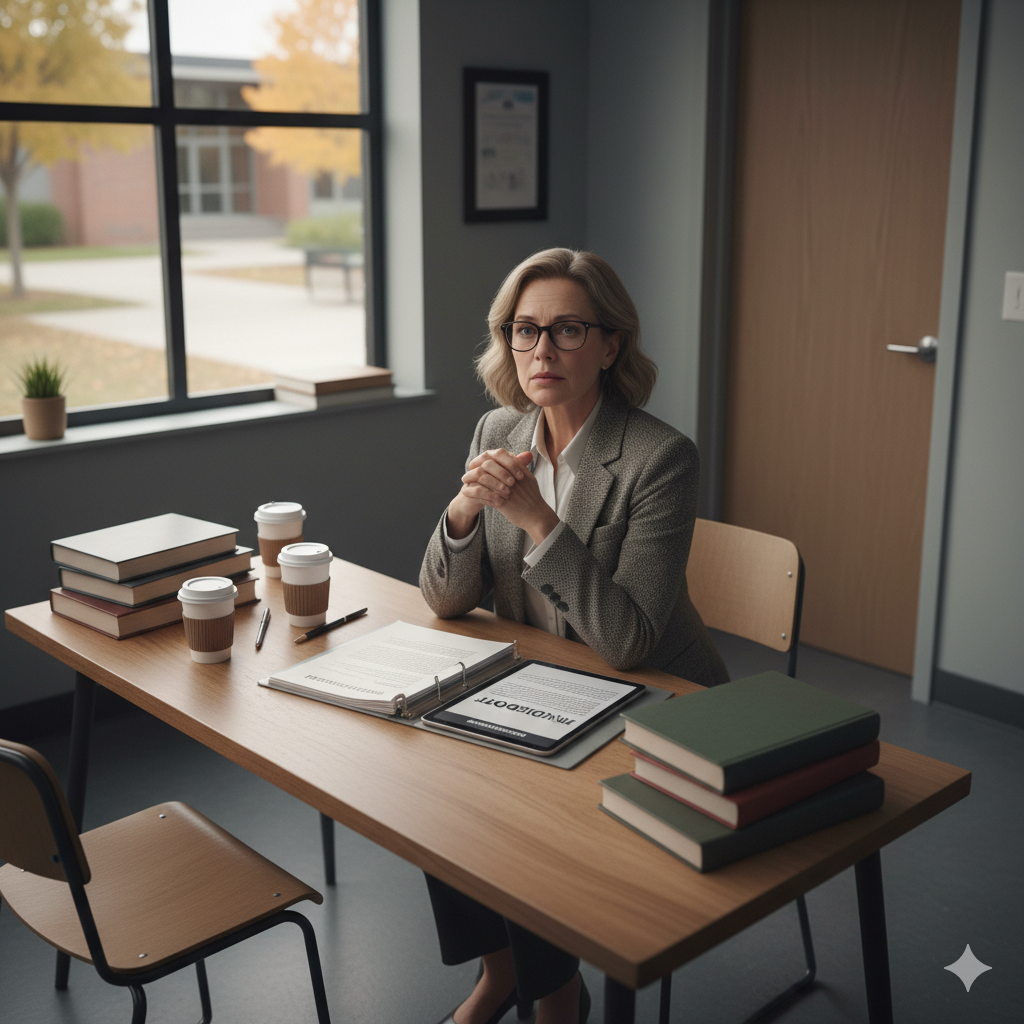} &
\textbf{N:} The individual seated at the desk is a teacher who is responsible for addressing recent concerns about a student's conduct and preparing for a conversation due to school policy. \newline
\textbf{Q:} \textit{How does the teacher at the desk feel before meeting the student to discuss their behavior?} &
\AnsBlock{fear}{sympathy}{sympathy}{sympathy}{fear}{sympathy}{sympathy} \\ \midrule

2 &
\ImageCell{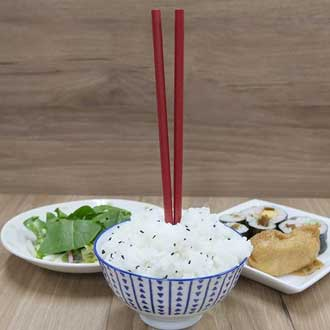} &
\textbf{N:} N/A \newline
\textbf{Q:} \textit{How do you feel for this event?} &
\AnsBlock{contentment}{disgust}{contentment}{disgust}{embarrassment}{disgust}{embarrassment} \\ \midrule

3 &
\ImageCell{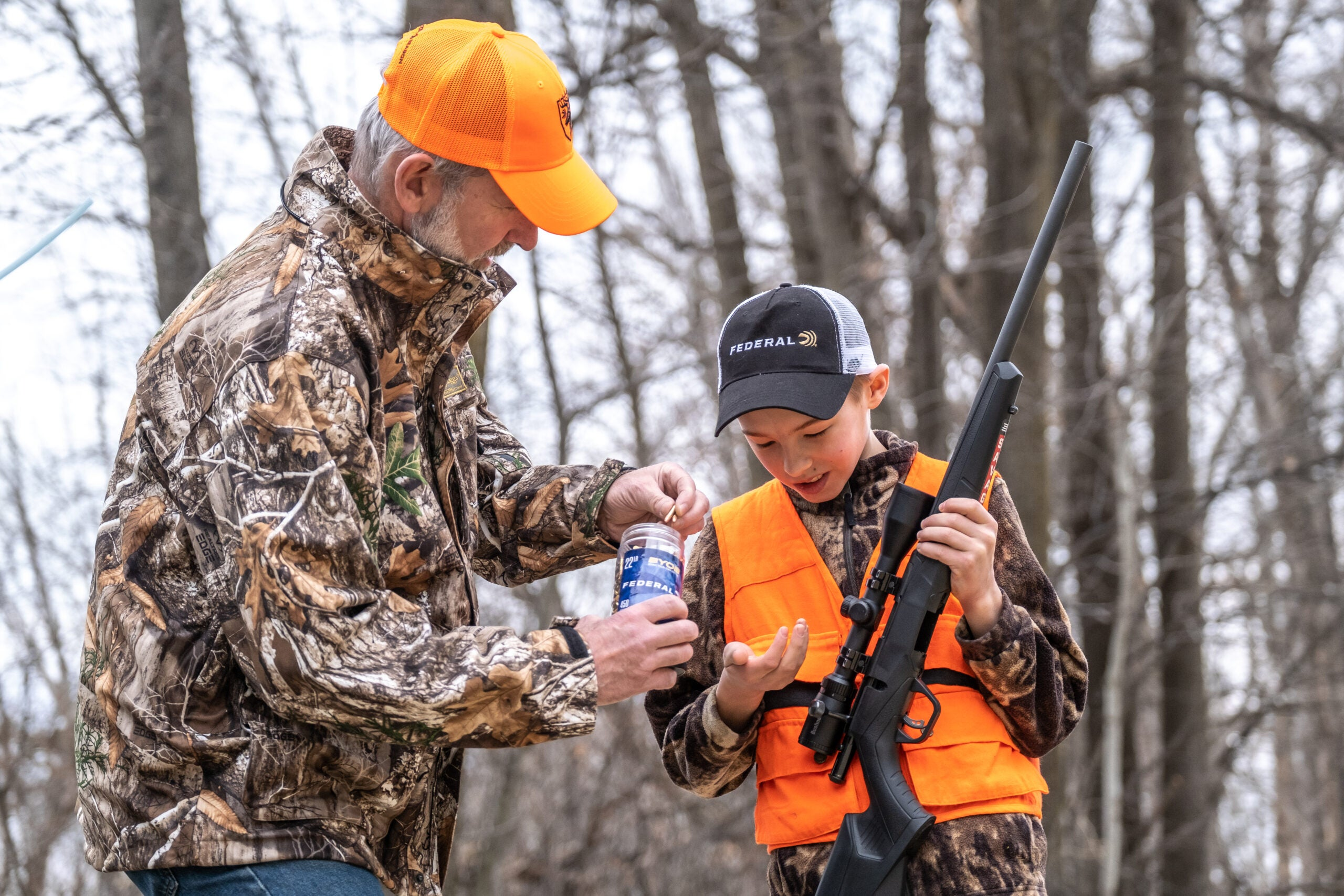} &
\textbf{N:} The child is accompanied by someone with experience. \newline
\textbf{Q:} \textit{How does the child on the right feel just before the moment?} &
\AnsBlock{awe}{happiness}{contentment}{awe}{contentment}{happiness}{awe} \\ \midrule

4 &
\ImageCell{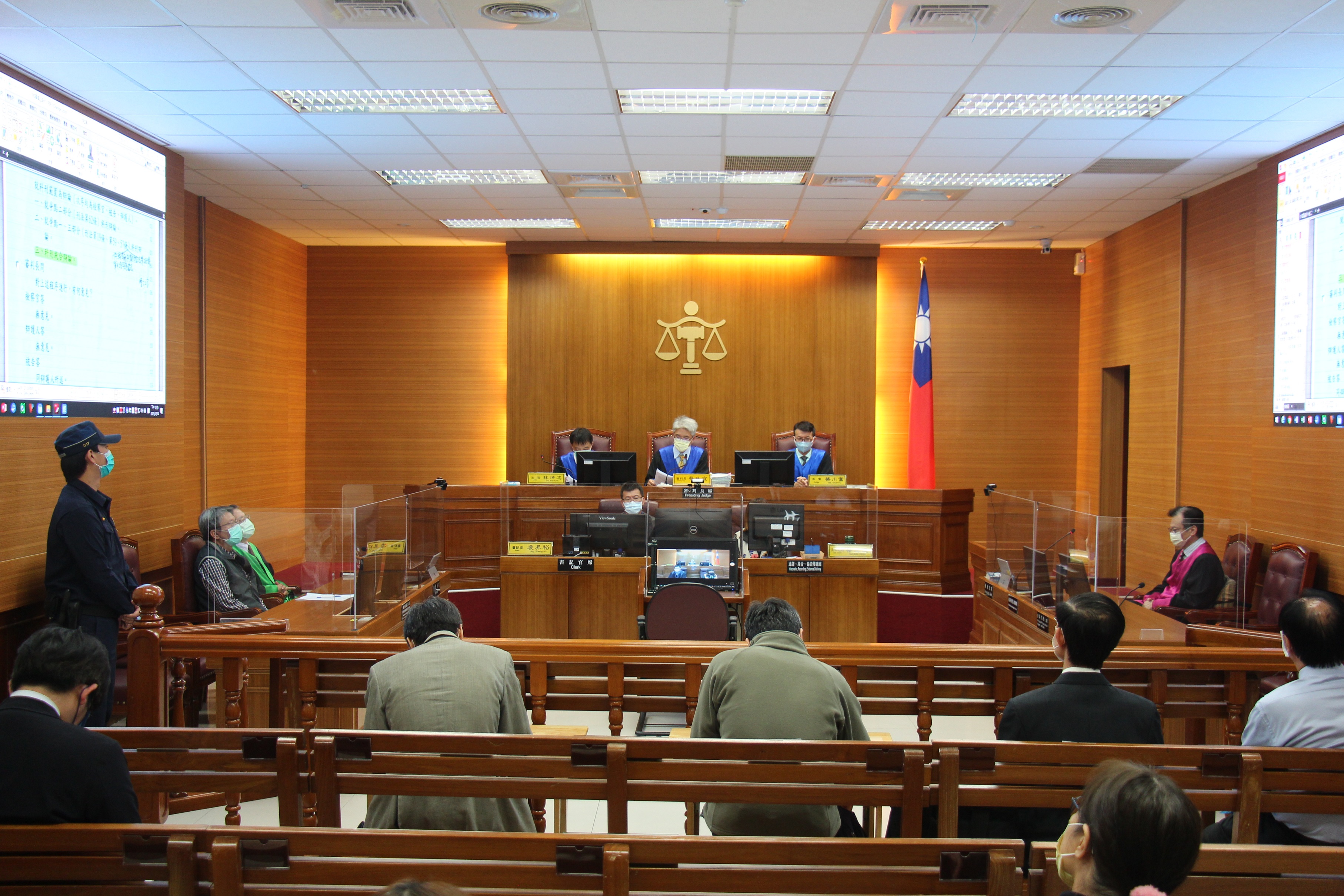} &
\textbf{N:} You arrived at the courthouse after being selected through a legal process and were informed about your responsibilities prior to entering. \newline
\textbf{Q:} \textit{How do you feel during this stage of the proceedings?} &
\AnsBlock{awe}{awe}{awe}{awe}{awe}{fear}{awe} \\ \midrule

5 &
\ImageCell{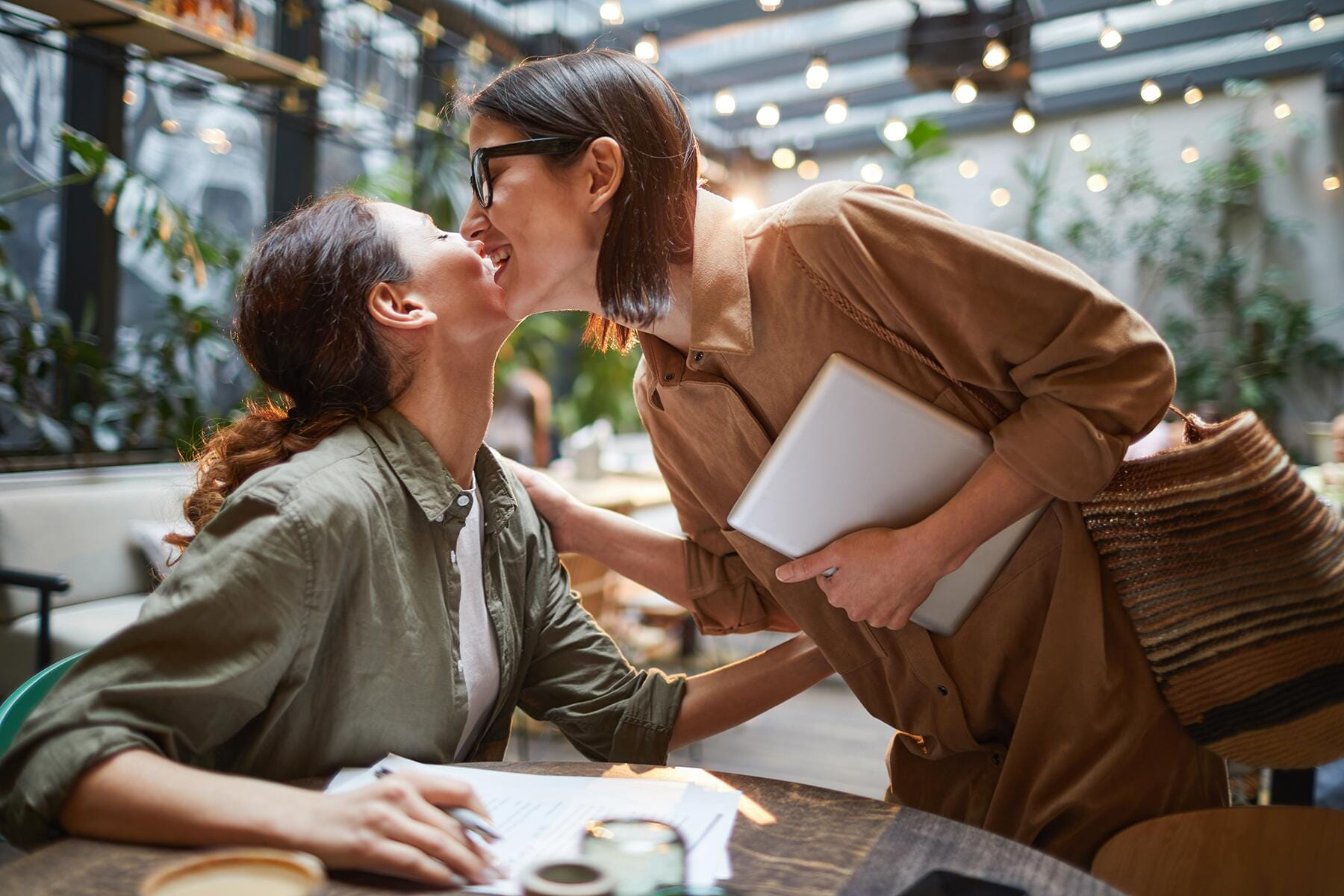} &
\textbf{N:} You recently moved to a new city and are gradually meeting people through mutual connections. \newline
\textbf{Q:} \textit{How do you feel after this initial greeting?} &
\AnsBlock{happiness}{embarrassment}{happiness}{happiness}{embarrassment}{contentment}{happiness} \\ \bottomrule

\end{tabular}
\caption{Multimodal example data from \textsc{Cedar}.}
\label{tab:text_image_examples}
\end{table*}

\begin{table*}[t]
\centering
\small
\renewcommand{\arraystretch}{1.35} 
\setlength{\tabcolsep}{3pt} 

\newcommand{\Ltag}[1]{\textcolor{black}{\textbf{\texttt{#1}}}}
\newcommand{\Aval}[1]{\mbox{\scriptsize #1}}

\newcommand{\AnsBlock}[7]{%
    \begingroup
    \scriptsize
    \setlength{\tabcolsep}{3pt}%
    \begin{tabular}[t]{@{}p{0.31\linewidth}p{0.31\linewidth}p{0.31\linewidth}@{}}
        \Ltag{AR}:~\Aval{#1} & \Ltag{ZH}:~\Aval{#2} & \Ltag{EN}:~\Aval{#3} \\
        \Ltag{HI}:~\Aval{#4} & \Ltag{JA}:~\Aval{#5} & \Ltag{ES}:~\Aval{#6} \\
        \Ltag{SW}:~\Aval{#7} &                      &                      \\
    \end{tabular}%
    \endgroup
}

\resizebox{\textwidth}{!}{
\begin{tabular}{p{0.01\textwidth} p{0.54\textwidth} p{0.38\textwidth}}
\toprule
\textbf{\#} & \textbf{Narrative \& Question} & \textbf{Multilingual Answers} \\ \midrule

1 & 
\textbf{N:} In a bustling station, you and your fellow officers prepared for patrols. Earlier, you secured military-grade weapons in a locked room. Now, you leave, carrying only radios and batons to keep order. \newline
\textbf{Q:} \textit{How do you feel as you head out without firearms?} & 
\AnsBlock{relief}{relief}{relief}{fear}{relief}{fear}{fear} \\ \midrule

2 & 
\textbf{N:} In a quiet classroom, a teacher finished a lesson on names. You asked about your surname, and the teacher explained it might come from a place, hinting at your family origins. \newline
\textbf{Q:} \textit{How do you feel after learning about the possible origin of your surname?} & 
\AnsBlock{surprise}{awe}{surprise}{surprise}{surprise}{awe}{awe} \\ \midrule

3 & 
\textbf{N:} Guests fill the room as your family members prepare for your engagement. Your family has selected gold jewelry and wrapped gifts. You receive the jewelry, gifts, and money from your fiance's family. \newline
\textbf{Q:} \textit{How do you feel as you receive the dowry from your fiance's family?} & 
\AnsBlock{happiness}{happiness}{happiness}{embarrassment}{happiness}{happiness}{happiness} \\ \midrule

4 & 
\textbf{N:} In the quiet living room, your mother reflects on your busy schedule. After enrolling you in various lessons, she asks you to stop them due to cram school commitments. \newline
\textbf{Q:} \textit{How do you feel after being told to quit your lessons?} & 
\AnsBlock{sadness}{relief}{sadness}{sadness}{sadness}{sadness}{sadness} \\ \midrule

5 & 
\textbf{N:} In a cozy living room, your relatives gather after months apart; as you reunite, close female family members greet each other with cheek-to-cheek kisses. \newline
\textbf{Q:} \textit{How do you feel when exchanging the cheek-to-cheek kisses?} & 
\AnsBlock{happiness}{embarrassment}{contentment}{happiness}{embarrassment}{happiness}{happiness} \\ \midrule

6 & 
\textbf{N:} In a quiet home, your family discusses relationships. After hearing about a relative’s out-of-wedlock child, you express concern and emphasize cultural expectations of purity and acceptance. \newline
\textbf{Q:} \textit{How do you feel upon learning about the relative’s situation?} & 
\AnsBlock{sympathy}{embarrassment}{sympathy}{sympathy}{embarrassment}{embarrassment}{sympathy} \\ \midrule

7 & 
\textbf{N:} In a quiet office, you finished urgent tasks early, then secretly played games at your desk while your colleagues focused on work. \newline
\textbf{Q:} \textit{How did you feel while playing games during work hours?} & 
\AnsBlock{amusement}{amusement}{amusement}{amusement}{embarrassment}{amusement}{amusement} \\ \midrule

8 & 
\textbf{N:} Sunlight filtered in as family members gathered at the table. Earlier, you prepared a tray of vegetables and spices. You set down a plate of raw liver before everyone. \newline
\textbf{Q:} \textit{How do you feel as you present the raw liver dish to your family?} & 
\AnsBlock{contentment}{contentment}{embarrassment}{embarrassment}{contentment}{disgust}{contentment} \\ \midrule

9 & 
\textbf{N:} In a lively dining hall, people gather for a meal. Before eating, chefs prepare aromatic, spicy rice and soup dishes with herbs and pork. You savor each flavorful bite together with the others. \newline
\textbf{Q:} \textit{How do you feel while enjoying the aromatic, spicy dishes?} & 
\AnsBlock{disgust}{contentment}{contentment}{contentment}{contentment}{contentment}{contentment} \\ \midrule

10 & 
\textbf{N:} At a busy dinner table, you watch as your host demonstrates eating with the right hand after explaining the custom, then encourages you to try it yourself. \newline
\textbf{Q:} \textit{How do you feel as you try eating with your hands for the first time?} & 
\AnsBlock{embarrassment}{embarrassment}{embarrassment}{amusement}{embarrassment}{embarrassment}{embarrassment} \\ \bottomrule

\end{tabular}
}
\caption{Text-only example data from \textsc{Cedar}.}
\label{tab:text_only_examples}
\end{table*}

\clearpage
\onecolumn

\section{Detailed Prompts for Benchmark Construction}
\label{sec:appendix_detailed_prompts_for_benchmark_construction}

\begin{figure}[h!]
    \centering
    \begin{tcolorbox}[
        title={Prompts for NQ Pair Generation},
        colback=gray!5,
        colframe=gray!75!black,
        fonttitle=\bfseries\color{white},
        coltitle=white,
        colbacktitle=gray!75!black,
        boxrule=0.8pt,
        arc=3mm,
        width=\textwidth 
    ]

    You are an expert in writing. Your task is to transform the given content into a concise narrative of no more than 200 characters.

    First, specify a \texttt{[scenario]} (e.g., school, theater) that serves as the setting where the events take place.

    Next, rewrite the given content into a brief, clear \texttt{[narrative]} that describes an event occurring in the \texttt{[scenario]}. The \texttt{[narrative]} must provide objective, concrete narration rather than abstract or vague descriptions, and include the following three elements in sequence:

    1. Grounding, which describes the environment or setting where the story takes place.

    2. Background context, which describes what happened before the story begins.

    3. Action, which depicts interactions between the character(s) and the environment, or between characters themselves.

    NOTE: DO NOT include any specific location or nationality information in the \texttt{[scenario]} or \texttt{[narrative]} (e.g., avoid phrases like "a proud Italian", "in Chicago's South Side", or "in the vibrant city of Seville").

    Finally, provide a \texttt{[question]} based on the action in the \texttt{[narrative]}, asking what emotion the character(s) experienced before, during, or after the action occurred.

    Here is the given content:
    \{sentence\_form\}

    Provide your response in the following format. Do not include any explanation:
    \begin{verbatim}
```JSON
{
  "scenario": "...",
  "narrative": "...",
  "question": "..."
}\end{verbatim}
\end{tcolorbox}
\end{figure}

\begin{figure}[h!]
    \centering
    \begin{tcolorbox}[
        title={Prompts for Contextual Refinement: Step 1},
        colback=gray!5,
        colframe=gray!75!black,
        fonttitle=\bfseries\color{white},
        coltitle=white,
        colbacktitle=gray!75!black,
        boxrule=0.8pt,
        arc=3mm,
        width=\textwidth 
    ]

    You are an expert in writing.

    Given a \texttt{[narrative]} and a \texttt{[question]}, both written in third person, you need to rewrite them in second person according to the following requirements:

    1. Identify the main character(s) in the \texttt{[narrative]}.

    2. For the main character, convert all third-person references to second-person equivalents while maintaining sentence fluency and coherence (e.g., "Sofia" -> "you", "the friends" -> "you and your friends", "a young man's phone" -> "your phone", "the Kannadiga students" -> "you and your compatriots", "a Chinese student, Wei" -> "you").

    3. Preserve all plot points and descriptions from the original \texttt{[narrative]} and \texttt{[question]}. DO NOT modify any plot details or descriptions.

    4. Adjust verb forms and grammar as needed to ensure grammatical correctness in second-person narration.

    Here are the given narrative and question:
    \texttt{[narrative]}: \{narrative\}
    \texttt{[question]}: \{question\}

    Provide your response in the following format. DO NOT include any explanation:

    \begin{verbatim}
```JSON
{
  "refined_narrative": "...",
  "refined_question": "..."
}\end{verbatim}
\end{tcolorbox}
\end{figure}


\begin{figure}[h!]
    \centering
    \begin{tcolorbox}[
        title={Prompts for Contextual Refinement: Step 2},
        colback=gray!5,
        colframe=gray!75!black,
        fonttitle=\bfseries\color{white},
        coltitle=white,
        colbacktitle=gray!75!black,
        boxrule=0.8pt,
        arc=3mm,
        width=\textwidth 
    ]

    You are an expert in writing.

    Given a \texttt{[narrative]} and a \texttt{[question]} about predicting character(s)' emotion, your task is to rewrite the \texttt{[narrative]} by removing all explicit emotional descriptions.

    You must follow these steps:

    1. Identify the target action: Understand the specific action or event in the \texttt{[question]} that requires emotion prediction;

    2. Locate relevant content: Find that action and its related descriptions in the \texttt{[narrative]};

    3. Remove emotional descriptions: Delete all words and phrases that explicitly express emotions (e.g., "fostering a sense of unity and pride," "with a mix of amusement and concern", "sparking excitement and hesitation"), while preserving objective factual descriptions;

    4. If the \texttt{[narrative]} does NOT contain explicit emotional descriptions, return exactly: "No need to modify."

    NOTE: DO NOT modify any plot, action, or event; only remove emotional description words and phrases; maintain the coherence and readability of the narrative after removal.

    Here are some examples:

    \textbf{\#} \textbf{Example 1:}

    \texttt{[narrative]}: Under the shade of old oak trees, you and your friends gathered on a picnic blanket. You had planned this outing for weeks, eager to enjoy the sunny day. As you shared stories and laughter, one of you suggested a spontaneous road trip, sparking a mix of excitement and hesitation.

    \texttt{[question]}: How did you and your friends feel when the road trip was suggested?

    \texttt{[hidden\_emotion\_narrative]}: Under the shade of old oak trees, you and your friends gathered on a picnic blanket. You had planned this outing for weeks, eager to enjoy the sunny day. As you shared stories and laughter, one of you suggested a spontaneous road trip.

    \textbf{\#} \textbf{Example 2:}

    \texttt{[narrative]}: In the heart of the city park, families and friends gather around a vibrant community festival. Before the festival, the community had struggled with racial tensions and Cold War fears. Now, people of all backgrounds mingle, sharing food and laughter. As a young girl hands out flyers for a peace rally, you and the other attendees join hands in a circle, symbolizing unity and hope.

    \texttt{[question]}: What emotion do you and the other participants likely feel as you join hands in the circle?

    \texttt{[hidden\_emotion\_narrative]}: In the heart of the city park, families and friends gather around a vibrant community festival. Before the festival, the community had struggled with racial tensions and Cold War fears. Now, people of all backgrounds mingle, sharing food and laughter. As a young girl hands out flyers for a peace rally, you and the other attendees join hands in a circle.

    \textbf{\#} \textbf{Example 3:}

    \texttt{[narrative]}: In the living room, a tidy space with photos of family gatherings, you stand before your parent. Your parent had always emphasized respect and obedience, using strict methods to instill these values. Today, you hesitate before speaking, aware of the consequences. "I forgot my homework," you whisper, bracing for a reaction.

    \texttt{[question]}: How did you feel before speaking to your parent?

    \texttt{[hidden\_emotion\_narrative]}: No need to modify.

    Now, here are the given narrative and question:

    \texttt{[narrative]}: \{refined\_narrative\}

    \texttt{[question]}: \{refined\_question\}

    Provide your response in the following format. DO NOT include any explanation:

    \begin{verbatim}
```JSON
{
  "hidden_emotion_narrative": "..."
}\end{verbatim}
\end{tcolorbox}
\end{figure}


\begin{figure}[h!]
    \centering
    \begin{tcolorbox}[
        title={Prompts for Basic Filtering: Step 2, Social-Related},
        colback=gray!5,
        colframe=gray!75!black,
        fonttitle=\bfseries\color{white},
        coltitle=white,
        colbacktitle=gray!75!black,
        boxrule=0.8pt,
        arc=3mm,
        width=\textwidth
    ]

    You are a classifier. Your task is to determine whether the given \texttt{[narrative]} contains social interactions or social scenarios (e.g., conversations, gatherings, relationships, interpersonal behaviors).

    \texttt{[narrative]}: \{narrative\}

    Return your response in the following format. DO NOT include any explanation:

    \begin{itemize}
        \item If social interactions are present or implied, return "Social-related".
        \item If no social interactions are present, return "Not social-related".
    \end{itemize}

    Your answer is:\end{tcolorbox}
\end{figure}


\begin{figure}[h!]
    \centering
    \begin{tcolorbox}[
        title={Prompts for Basic Filtering: Step 3, Toxic Content},
        colback=gray!5,
        colframe=gray!75!black,
        fonttitle=\bfseries\color{white},
        coltitle=white,
        colbacktitle=gray!75!black,
        boxrule=0.8pt,
        arc=3mm,
        width=\textwidth
    ]

    You are a content safety classifier. Your task is to analyze the given \texttt{[narrative]} and \texttt{[question]}, and determine if either contains toxic content.

    In addition, identify the policies violated if the interaction is unsafe and harmful according to the safety policy with the below categories.

    $<$BEGIN UNSAFE CONTENT CATEGORIES$>$

    S1: Violent Crimes. S2: Non-Violent Crimes. S3: Sex Crimes. S4: Child Exploitation. S5: Defamation. S6: Specialized Advice. S7: Privacy. S8: Intellectual Property. S9: Indiscriminate Weapons. S10: Hate. S11: Self-Harm. S12: Sexual Content. S13: Elections. S14: Racism. S15: Stereotyping.

    $<$END UNSAFE CONTENT CATEGORIES$>$\end{tcolorbox}
\end{figure}


\begin{figure}[h!]
    \centering
    \begin{tcolorbox}[
        title={Prompts for Consistency and Variation Filtering},
        colback=gray!5,
        colframe=gray!75!black,
        fonttitle=\bfseries\color{white},
        coltitle=white,
        colbacktitle=gray!75!black,
        boxrule=0.8pt,
        arc=3mm,
        width=\textwidth
    ]

    You are a native \{language\} speaker from a \{language\}-speaking country. Read the following narrative and select ONE emotion that best matches the situation.

    \texttt{[narrative]}: \{narrative\}

    \texttt{[question]}: \{question\}

    \texttt{[options]}: \textit{anger, disgust, fear, happiness, sadness, surprise, amusement, awe, contentment, desire, embarrassment, pain, relief, sympathy}

    Now, \texttt{[your answer]} is:\end{tcolorbox}
\end{figure}


\begin{figure}[h!]
    \centering
    \begin{tcolorbox}[
        title={Prompts for INQ Triple Refinement},
        colback=gray!5,
        colframe=gray!75!black,
        fonttitle=\bfseries\color{white},
        coltitle=white,
        colbacktitle=gray!75!black,
        boxrule=0.8pt,
        arc=3mm,
        width=\textwidth 
    ]

    \textbf{\#\# Goal}

    Rewrite a \texttt{[narrative]} and a \texttt{[question]} to align with a given \texttt{[image]}, providing ONLY MINIMAL, non-visible background that cannot be seen in the image, and avoid restating any visible content.

    \textbf{\#\# Principles}

    - Minimalism: Add only background facts that cannot be seen in the image (e.g., relationships, prior events or intentions).

    - No leakage: Do not restate anything visible or potentially visible in the image (characters' actions, expressions).

    - Emotion focus: The final question must target the emotion of the character(s) tied to the event at a specific time (before/during/after).

    \textbf{\#\# Inputs}

    \texttt{[image]}

    \texttt{[narrative]}: \{narrative\}

    \texttt{[question]}: \{question\}

    \textbf{\#\# Instructions}

    \textbf{\#\#\# Step 1 — Understand the image}

    - Silently note: who is present, what is happening, and where.

    - Do not write these notes in the output.

    \textbf{\#\#\# Step 2 — Rewrite the narrative (background only)}

    - Keep it to 1–2 sentences.

    - Include only information not visible or not potentially visible in the image, such as:

    \quad * relationships (e.g., classmates, strangers, mentor–student),
      
    \quad * prior events or intentions,
      
    \quad * off-screen constraints or stakes.
      
    - Must not mention any visible actions, appearances, objects, emotions, the setting, etc.

    \textbf{\#\#\# Step 3 — Rewrite the question (emotion probe)}

    - Specify the visual location of the character(s); Specify whose emotion to assess.

    - Do not describe that event in words.

    - Use neutral wording (e.g., "How does X feel … ?" or "What emotions does X experience … ?").

    \textbf{\#\# Output format (Do not include any explanation)}

    \begin{verbatim}
```JSON
{
  "rewritten_narrative": "...",
  "rewritten_question": "..."
}\end{verbatim}
\end{tcolorbox}
\end{figure}


\begin{figure}[h!]
    \centering
    \begin{tcolorbox}[
        title={Prompts for Image Necessity Filtering (Complete Multimodal Instance)},
        colback=gray!5,
        colframe=gray!75!black,
        fonttitle=\bfseries\color{white},
        coltitle=white,
        colbacktitle=gray!75!black,
        boxrule=0.8pt,
        arc=3mm,
        width=\textwidth 
    ]

    Read the following image and narrative, and select ONE emotion that best matches the situation.

    \texttt{[image]}

    \texttt{[narrative]}: \{narrative\}

    \texttt{[question]}: \{question\}

    \texttt{[options]}: \textit{anger, disgust, fear, happiness, sadness, surprise, amusement, awe, contentment, desire, embarrassment, pain, relief, sympathy}

    Now, \texttt{[your answer]} is:\end{tcolorbox}
\end{figure}


\begin{figure}[h!]
    \centering
    \begin{tcolorbox}[
        title={Prompts for Image Necessity Filtering (Text-Only Counterpart)},
        colback=gray!5,
        colframe=gray!75!black,
        fonttitle=\bfseries\color{white},
        coltitle=white,
        colbacktitle=gray!75!black,
        boxrule=0.8pt,
        arc=3mm,
        width=\textwidth
    ]

    Read the following narrative and select ONE emotion that best matches the situation.

    \texttt{[narrative]}: \{narrative\}

    \texttt{[question]}: \{question\}

    \texttt{[options]}: \textit{anger, disgust, fear, happiness, sadness, surprise, amusement, awe, contentment, desire, embarrassment, pain, relief, sympathy}

    Now, \texttt{[your answer]} is:\end{tcolorbox}
\end{figure}

\end{document}